%% file: main.tex
\documentclass[letterpaper]{article} 
\usepackage{arxiv}  
\usepackage{times}  
\usepackage{helvet} 
\usepackage{courier} 
\usepackage[hyphens]{url} 
\usepackage{graphicx}
\usepackage{natbib}  
\usepackage{caption} 
\usepackage{algorithm}
\usepackage{algorithmic}
\usepackage{newfloat}
\usepackage{listings}

\DeclareCaptionStyle{ruled}{labelfont=normalfont,labelsep=colon,strut=off}
\floatstyle{ruled}
\newfloat{listing}{tb}{lst}{}
\floatname{listing}{Listing}

\usepackage{graphicx}
\usepackage{multirow}
\usepackage{booktabs}
\usepackage{amsfonts}
\usepackage{amssymb}
\usepackage{tikz}
\usepackage{siunitx}
\usepackage{amsmath}
\usepackage{amsthm}
\usepackage{subcaption}
\usepackage{todonotes}

\title{Enhancing Chemical Explainability Through Counterfactual Masking}

\author{
    \L{}ukasz Janisi\'{o}w\textsuperscript{\rm 1,2,4}, 
    Marek Kocha\'{n}czyk\textsuperscript{\rm 1,$\dag$}, Bartosz Zieli\'{n}ski\textsuperscript{\rm 1}, Tomasz Danel\textsuperscript{\rm 1,3,$\ast$}
}
\affiliations{
    \textsuperscript{\rm 1} Faculty of Mathematics and Computer Science, Jagiellonian University, Krak\'{o}w 30-348, Poland\\
    \textsuperscript{\rm 2} Doctoral School of Exact and Natural Sciences, Jagiellonian University, Krak\'{o}w 30-348, Poland\\
    \textsuperscript{\rm 3} Faculty of Chemistry, Jagiellonian University, Krak\'{o}w 30-387, Poland\\
    \textsuperscript{\rm 4} Faculty of Economic Sciences, University of Warsaw, Warsaw 00-241, Poland\\
    
    \textsuperscript{$\ast$} tomasz.danel@uj.edu.pl
}

\begin{document}

\maketitle

\begin{abstract}
Molecular property prediction is a crucial task that guides the design of new compounds, including drugs and materials. While explainable artificial intelligence methods aim to scrutinize model predictions by identifying influential molecular substructures, many existing approaches rely on masking strategies that remove either atoms or atom-level features to assess importance via fidelity metrics. These methods, however, often fail to adhere to the underlying molecular distribution and thus yield unintuitive explanations. In this work, we propose counterfactual masking, a novel framework that replaces masked substructures with chemically reasonable fragments sampled from generative models trained to complete molecular graphs. Rather than evaluating masked predictions against implausible zeroed-out baselines, we assess them relative to counterfactual molecules drawn from the data distribution. Our method offers two key benefits: (1) molecular realism underpinning robust and distribution-consistent explanations, and (2) meaningful counterfactuals that directly indicate how structural modifications may affect predicted properties. We demonstrate that counterfactual masking is well-suited for benchmarking model explainers and yields more actionable insights across multiple datasets and property prediction tasks. Our approach bridges the gap between explainability and molecular design, offering a principled and generative path toward explainable machine learning in chemistry.
\end{abstract}

\section{Introduction}

Molecular property prediction has emerged as a cornerstone of modern drug discovery and materials science, promising to dramatically accelerate the identification of compounds with desired characteristics. By leveraging machine learning algorithms to predict properties such as binding affinity, toxicity, solubility, and biological activity, researchers can efficiently navigate the vast chemical space without exhaustive experimental testing \citep{wieder2020property, dara2022machine}. However, the increasing complexity of state-of-the-art models presents a significant challenge: while deep neural networks and ensemble methods deliver impressive predictive performance, they often function as inscrutable ``black boxes,'' offering little insight into the structural features driving their predictions \citep{biecek2024position, longo2024xai}. This opacity is particularly problematic in chemistry and pharmaceutical development, where understanding structure--property relationships is essential for rational molecular design, regulatory approval, and scientific knowledge advancement. Interpretable models that can explain their predictions in chemically meaningful terms are, therefore, not merely desirable but necessary to bridge the gap between statistical performance and actionable scientific insights \citep{luna2020drug, wong2024discovery}.

\begin{figure*}
    \centering
    \includegraphics[width=0.8\linewidth]{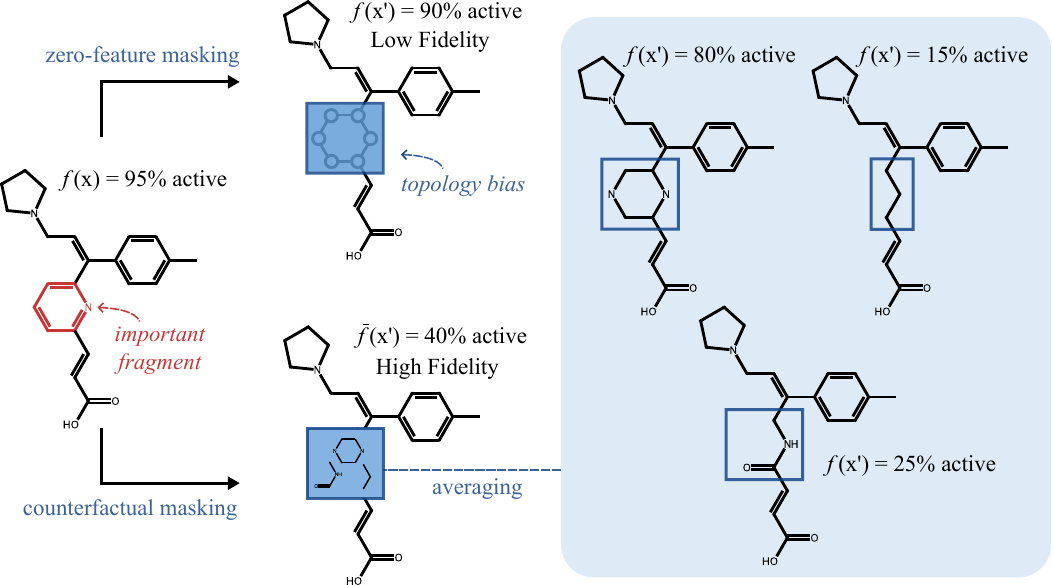}
    \caption{Comparison between zero-feature masking and our CM method. The former strategy preserves the topology of the masked fragment, leading to a potential predictive bias. CM generates multiple fragment replacements, enabling more robust and trustworthy evaluation of model explanations.}
    \label{fig:abstract}
\end{figure*}

Graph masking has emerged as a fundamental technique for explaining molecular property predictions, serving both as the foundation for certain explainable AI (XAI) methods and as the basis for evaluation metrics like fidelity \citep{bugueno2024graph}. By systematically removing or obscuring specific atoms or substructures, masking approaches attempt to quantify each component's contribution to the final prediction. However, conventional masking strategies suffer from critical limitations in the molecular domain. When atoms or bonds are naively masked, the resulting structures often become chemically implausible or physically impossible, generating examples that fall outside the training distribution. This out-of-distribution problem undermines the reliability of both the explanations and their evaluation metrics. Furthermore, current masking approaches inadvertently leak information about the original graph topology to the model—even when certain atoms are ``masked,'' their structural relationships and connectivity patterns remain implicitly encoded in the modified graph. This information leakage creates a false sense of explanation quality, as models may still leverage these implicit structural cues rather than truly operating without knowledge of the masked components. Developing more sophisticated masking techniques that preserve chemical validity while effectively controlling information flow is therefore essential for advancing trustworthy explanations in molecular machine learning.

Current approaches to molecular property prediction primarily rely on \textit{post-hoc} explanation methods such as gradient-based attribution, attention mechanisms, and feature importance scores \citep{pope2019explainability}. These techniques attempt to highlight atoms or substructures that significantly influence predictions, providing chemists with visual maps of ``important'' molecular regions. However, conventional explanation methods struggle to address the holistic nature of molecular design, where virtually all components contribute to functionality through precise atom arrangements. Identifying which atoms are ``important'' offers limited insight when the entire structure has been carefully optimized. What chemists truly need is to understand why these structural elements matter and how they might be modified to achieve the desired properties. This gap between highlighting important features and providing actionable insights motivates the development of counterfactual explanations, which instead focus on answering the critical question of what other molecular design choices would lead to a change in chemical properties, e.g., higher potency or lower toxicity \citep{gleeson2011potency, guengerich2011tox}.

In this paper, we present counterfactual masking (CM) for molecular graphs, a novel method designed to enhance the explanations of machine learning model predictions (Figure~\ref{fig:abstract}). For every key molecular fragment identified with the explanation method, we generate a set of counterfactual explanations that replace the original fragment. This visually illustrates the importance of these fragments in relation to specific properties. Our approach aims to provide clearer insights into why certain molecular features contribute to predictive outcomes. The contribution of this paper can be summarized as follows:
\begin{enumerate}
    \item We introduce a new masking method for molecular graphs that hides molecular fragments by replacing them with a set of alternative generated fragments, ensuring that masked molecules are valid in-distribution molecules.
    \item We use our masking technique to provide additional insights for the explanation model by producing counterfactual explanations that show why these structures might be important for the model.
\end{enumerate}

The source code and accompanying data are available at \linebreak \url{https://github.com/gmum/counterfactual-masking}.

\section{Related Work}

\paragraph{Factual explanations.} Factual explanation methods for GNNs can be broadly categorized into gradient-based, perturbation-based, surrogate-based, and self-interpretable approaches. Gradient-based techniques, such as CAM~\citep{zhou2016learning}, Grad-CAM~\citep{selvaraju2017grad}, and Integrated Gradients~\citep{sundararajan2017axiomatic}, estimate feature relevance through backpropagation. Perturbation-based methods, including GNNExplainer~\citep{ying2019gnnexplainer} and SubgraphX~\citep{yuan2021explainability}, generate explanations by learning masks or searching over subgraphs that are most influential to model predictions. In contrast, self-interpretable GNNs attempt to provide inherently explainable reasoning by embedding interpretability directly into the architecture~\citep{zhang2022protgnn}. While factual methods assign importance scores to nodes, edges, or subgraphs, they often lack mechanisms to verify the factual correctness of the highlighted structures on real-world datasets. Moreover, most approaches offer limited insight into why a particular structure is deemed important.

\paragraph{Counterfactual explanations.} Unlike factual methods, counterfactual explanations provide additional insight into model predictions by showing similar instances that are predicted differently from the original. In chemistry, research on counterfactual methods is still limited compared to other fields, but several methods have been proposed. \citet{wellawatte2022model} introduced MMACE, a model-agnostic approach that generates counterfactual examples by insertions, replacements, or deletions in the SELFIES~\citep{krenn2020self} representation of the compound. \citet{numeroso2021meg} proposed MEG, a method employing reinforcement learning to find minimal atom-level modifications that change the model prediction. Other methods tweak the latent representation to sample close analogs of a compound, e.g. by using a variational autoencoder~\citep{wang2024global}. However, many counterfactual methods are difficult to apply due to the discrete nature of molecular graphs, and some of the methods introduce modifications that hurt synthetic accessibility and lack localization of the changes.

\paragraph{Generative models for structure optimization.} Our method generates replacements only in the parts of the molecule that are predicted as important. Some methods have been proposed for such context-constrained generative modeling. For example, models such as DiffLinker~\citep{igashov2024equivariant} and DeLinker~\citep{imrie2020deep} were proposed to generate fragments between two or more molecular fragments. \citet{polishchuk2020crem} introduced CReM, a method that finds fragments in compound databases that were already seen in the same molecular context, ensuring the synthesizability of the compounds generated. Recently, \citet{lee2025genmol} proposed GenMol, a textual masked discrete diffusion model capable of generating linkers and scaffold decorations.

\section{Methods}

Let us define a molecular graph as $\mathcal{G}=(\mathcal{V}, \mathcal{E}, F)$, where $\mathcal{V}$ is a set of nodes, $\mathcal{E}$ is a set of edges, and $F:\mathcal{V}\to \mathbb{R}^d$ is a function that assigns features to nodes. Factual explanation techniques for molecular graphs identify the atoms that significantly affect model predictions. This can be formally described by a function $E:\mathcal{V}\to \{0,1\}$ that assigns a value of one to nodes considered crucial for the prediction. Next, we will present our CM method, which evaluates the influence of these important fragments and offers alternative subgraphs for complementary counterfactual explanations.

\subsection{Counterfactual Masking}

CM generates a set of alternative molecules by replacing fragments identified by a factual explanation technique. 

\paragraph{Step 1: Identification of important subgraphs.} All important connected subgraphs are extracted. First, all the nodes crucial for the prediction are identified using a factual explanation method, resulting in a set of key nodes $\mathcal{V}_\mathrm{imp}=\{v: v\in\mathcal{V} \land E(v)=1\}$. Next, all these key nodes and their connected edges are removed, and the remaining graph serves as a context $\mathcal{C}$ for the generative methods to fill in the missing information. The context should also include attachment points $\mathcal{A}$, which are nodes from $\mathcal{C}$ that were connected to any of the removed key nodes.

\paragraph{Step 2: Regeneration of the removed fragments.} A generative model $g$ is employed to replace each removed subgraph based on the context and the attachment points, modeling the distribution of feasible molecules $p(x\,|\,\mathcal{C}, \mathcal{A})$.
Then, CM is a set of molecules produced by drawing multiple samples from the generator.

\paragraph{Properties of our masking method.} Since CM is sampled from a distribution of feasible molecules, we can evaluate the quality of factual explanations by comparing the original prediction with the average case from that distribution rather than using artificially zeroed-out features. In the case of classification, we can present molecules that are predicted to be in a different class as counterfactual examples.

\paragraph{Generative model.} As a generative model for fragment replacement, we use two different approaches. One is based on the CReM algorithm, which uses fragments extracted from the ChEMBL database. In this approach, only fragments that were already seen in the same atom context can be used, thus increasing the synthetic accessibility of the generated replacements. In this case, the entire molecular ring is replaced if any atom within it is found to be influential because CReM does not operate on partial rings.

Another generative method used is DiffLinker, a diffusion model that produces a linker between fragments located in 3-D space, trained on the ZINC dataset~\citep{irwin2005zinc}. In this method, the input molecules are embedded in 3-D space by applying a force field method to generate a stable conformation of a molecule. This ensures that the generated fragment replacement occupies a similar amount of space between fragments being connected. A considerable disadvantage of this method is that the diffusion model works at the atomic level, sometimes resulting in unsynthesizable or even invalid molecules.

\subsection{Common Substructure Pair Dataset}

To evaluate the performance of masking methods, we introduced a new dataset. Starting with the Solubility dataset \citep{sorkun2019aqsoldb} from the Therapeutics Data Commons (TDC) collection \citep{huang2021tdc}, we applied BRICS decomposition \citep{degen2008BRICS} to fragment each molecule into chemically meaningful substructures. These substructures were then used to query the PubChem database \citep{kim2025pubchem} to identify larger molecules (superstructures) in which each substructure appears as a subgraph. For each source fragment, multiple matching superstructures were identified, resulting in a dataset of 881 pairs in the form (source fragment, superstructures). On average, each source fragment is associated with 52 chemically distinct superstructures.

\section{Results}

In this section, we first demonstrate that the currently widely used masking technique of zeroing features is flawed, and our masking strategy resolves this issue. Next, we show the utility of our method in producing counterfactual explanations. CM is then used to benchmark XAI techniques. We conclude by discussing the limitations of our method.

\subsection{Enhanced atom masking strategy}

\begin{table*}[tb]
  \caption{The effectiveness of masking methods in preventing information leakage from the masked subgraph. The difference of predictions ($|\Delta\hat{y}|$) should be close to zero when only the common subgraph is retained, while the rest of the molecule is masked. Results are shown for three scenarios: single anchor (a non-shared fragment between the two molecules has one attachment point to the common substructure), multiple anchors (the fragment has more than one attachment point), and both (the fragment can have one or more attachment points). Examples of molecules for each scenario are provided in Appendix B.}
  \label{tab:pair_eval}
  \centering
  \begin{tabular}{lccccccccc}
    \toprule & \multicolumn{3}{c}{Single Anchor} & \multicolumn{3}{c}{Multiple Anchors} & \multicolumn{3}{c}{Both}                     \\
    \cmidrule(r){2-4} \cmidrule(lr){5-7} \cmidrule(l){8-10}
    Masking     & $|\Delta\hat{y}|$ $\downarrow$ & Validity $\uparrow$ & Size $\uparrow$ & $|\Delta\hat{y}|$ $\downarrow$ & Validity $\uparrow$ & Size $\uparrow$ & $|\Delta\hat{y}|$ $\downarrow$ & Validity $\uparrow$ & Size $\uparrow$ \\
    \midrule
    \textit{No masking} & 2.81 & 100\% & 517 & 2.52 & 100\% & 638 & 2.91 & 100\% & 783\\
    Feature zeroing & 2.08 & 100\% & 517 & 1.60 & 100\% & 638 & 1.80 & 100\% & 783\\
    CM (CReM)  & 1.68 & 88\% & 454 & 1.32 & 53\% & 341 & 1.72 & 65\% & 510\\
    CM (DiffLinker)  & 1.81 & 67\% & 347 & 1.59 & 61\% & 386 & 1.55 & 65\% & 508\\
    \bottomrule
  \end{tabular}
\end{table*}

\begin{figure*}[tb]
  \centering
  \includegraphics[width=0.85\linewidth]{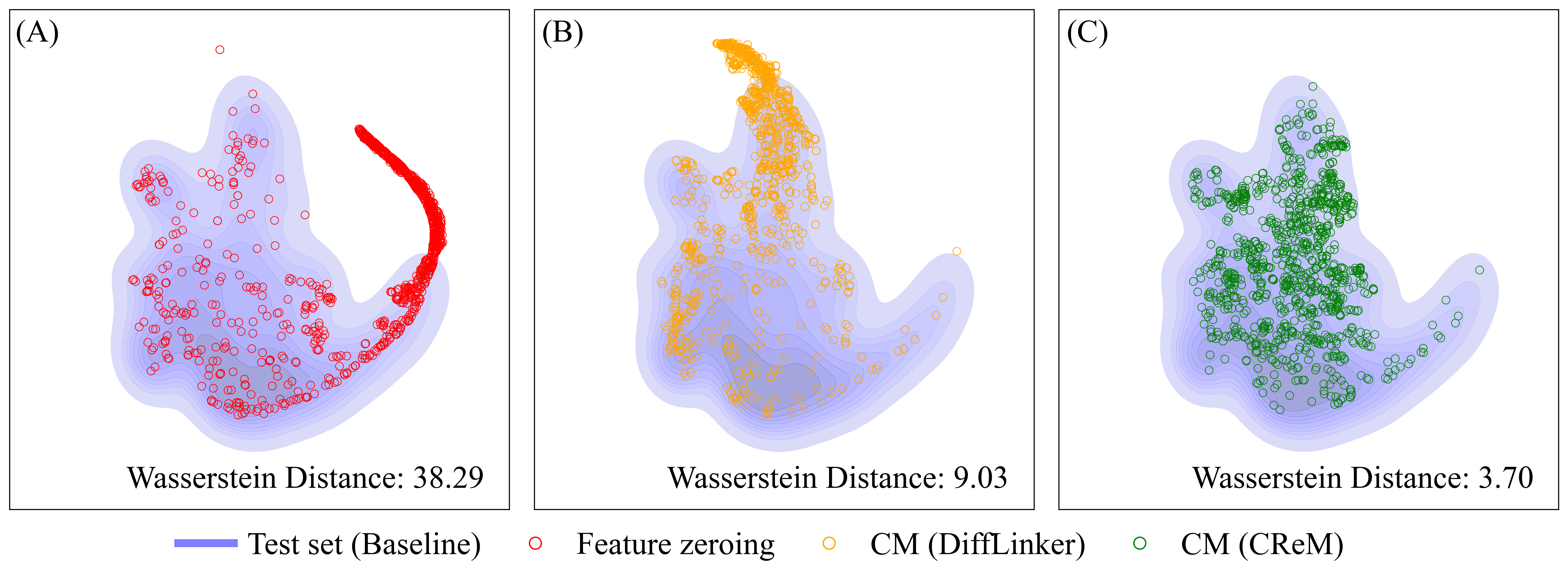}
  \caption{t-SNE visualization of molecular embeddings comparing the test set distribution with molecules containing parts masked by various methods. Wasserstein distances quantify the distributional divergence from the test set reference distribution.}
  \label{fig:embedding_comparison_testSet}
\end{figure*}

Masking is commonly used for creating or evaluating graph-based explanation methods. However, current masking techniques, such as feature zeroing, often produce samples outside the training distribution and may reveal information about the original graph topology. In this section, we examine our CM approach, where masked fragments are replaced with chemically plausible alternatives. We compare this method to the traditional feature zeroing baseline, where all node features within the masked fragments are set to zero.

\subsubsection{Setup.}

To assess the effectiveness of various masking techniques, we used the common substructure pair dataset described in the Methods section. For each substructure, we selected a pair of superstructures showing the greatest difference in solubility predictions from a GNN regression model described below. In each pair, we masked the molecular fragments that are not shared between the two molecules and measured the resulting difference in the model's predictions. An ideal masking technique should completely obscure the masked fragment, resulting in identical predictions for both molecules in the pair.

\subsubsection{Model.}

As the regression model, we used a Graph Isomorphism Network (GIN)~\citep{xu2018powerful} with three layers, each having a hidden size of 512. Global mean pooling was used to aggregate node features. The model was trained for 300 epochs with a batch size of 64 using the Adam optimizer and mean squared error (MSE) loss. During training, a dropout rate of 30\% was applied.

\subsubsection{Metrics.}

We used three metrics to evaluate the performance of each masking method. $\boldsymbol{|\Delta\hat{y}|}$ measures the absolute difference in predicted values between two molecules sharing the same substructure. \textbf{Validity} is the percentage of molecular pairs for which the method generates valid masked molecules, and \textbf{Size} is the total number of pairs the method can be successfully applied to.

\subsubsection{Results.}

Table~\ref{tab:pair_eval} compares the CM approach using two generators with the baseline that sets the features of masked nodes to zero. All masking methods reduce prediction differences compared to unmasked pairs, with CM consistently outperforming the baseline in all three scenarios. CReM sampling produces the lowest prediction difference in two scenarios, while DiffLinker performs best in the "Both" scenario. However, CM depends on generative models, which may not always produce valid fragments, whereas feature zeroing provides complete coverage of the dataset.

Figure~\ref{fig:embedding_comparison_testSet} displays the distribution of molecular embeddings from the single anchor part of the dataset (for other scenarios, see Appendix B) after masking, compared to the test set of the solubility dataset. Since the test set is randomly sampled, it reflects the training data distribution. Embeddings of molecules masked with CM align more closely with this training distribution than those masked with feature zeroing, as evidenced by lower Wasserstein distances, especially for the CReM model. Molecules masked by feature zeroing often fall outside the training distribution, making their predictions and explanations less reliable. In contrast, CM provides a more robust and chemically grounded comparison by contrasting the original molecule’s prediction with the average prediction for chemically plausible alternatives. This method provides a meaningful measure of the influence of masked molecular fragments.

\subsection{Counterfactual explanations}

Although originally designed for masking tasks, CM can effectively generate realistic, chemically valid counterfactual examples by ensuring molecules stay within the data distribution. It also enables targeted local modifications to specific fragments, improving interpretability by offering insights into how substructure changes affect predicted classes.

\subsubsection{Dataset.}

To evaluate counterfactual explanations, we used three binary classification datasets from TDC: prediction of inhibition of cytochrome P450 enzymes CYP3A4 and CYP2D6, and hERG channel blockage (hERG), a key cardiotoxicity indicator \cite{karim2021herg}. We trained a GIN classification model using 80\% of each dataset, while the remaining 20\% was reserved for counterfactual generation.

\subsubsection{Explained model.}

To predict molecular properties, we used a GIN model with three layers, each with a hidden size of 512. Global mean pooling was applied to aggregate node features. The model was trained for 300 epochs with a batch size of 16, using the Adam optimizer and binary cross-entropy loss. A dropout rate of 30\% was applied during training, and early stopping with a patience of 20 epochs was used if no improvement was observed on the validation set.

\subsubsection{Counterfactual models and baselines.}

To generate counterfactuals using the CM approach, we first applied Grad-CAM to identify the top 20\% most important atoms in the molecule. These atoms were then replaced with new fragments using either the CReM or the DiffLinker framework to produce counterfactual examples. After generation, post-hoc filtering was applied to select alternative molecules that successfully changed the prediction class. From these, a subset was selected to maximize similarity to the original molecule while maintaining diversity among the already selected counterfactuals. The first example was chosen based on its highest similarity to the original sample, and subsequent counterfactuals were selected by balancing similarity to the original with diversity relative to those already chosen.

As baseline methods for counterfactual generation, we used GNNExplainer and Nearest Neighbor (NN). For GNNExplainer, counterfactuals were generated by removing the top 10\% of nodes identified as the most influential. For the NN method, counterfactuals were defined as the most structurally similar molecules from the training set that belong to the opposite prediction class. Additionally, we benchmarked our approach against MMACE, which produces counterfactuals by performing up to two targeted structural modifications (deletion, replacement, or insertion) to the original molecule.

\subsubsection{Metrics.}

We evaluated counterfactual generation using five metrics. \textbf{Before validity (BV)} measures the proportion of generated molecules that change the prediction class before any filtering is applied, while \textbf{validity} measures this proportion after filtering. \textbf{Similarity} indicates the average similarity between the original molecule and the filtered counterfactuals, calculated using the Tanimoto distance between molecular fingerprints. \textbf{Diversity} captures the variation among the generated counterfactuals, based on the Tanimoto distance between their fingerprints. Finally, \textbf{synthetic accessibility (SA)} evaluates the average ease of synthesizing the filtered counterfactuals.

\subsubsection{Results.}

The performance of counterfactual generation methods is summarized in Table~\ref{tab:counterfactual_explanations}. Before filtering, the NN baseline shows the highest validity, as it directly searches for molecules that change the prediction class. Among generative methods, CM (DiffLinker) achieves the highest rate of class changes in two out of three datasets. After filtering, all methods except GNNExplainer, which provides only one counterfactual per instance, reach a validity score of 100\%. MMACE generates the most similar and diverse molecules. However, CM (CReM) remains close in both staticstics across all tested datasets. The only exception is the NN baseline, which exhibits greater diversity than both because of its broader distance from the original molecules in chemical space. In terms of synthetic accessibility, CM (CReM) produces the most feasible molecules. Overall, combining Grad-CAM with CM (CReM) offers effective counterfactuals, on par with MMACE, but with improved synthetic accessibility.

\begin{table*}[tb]
    \caption{Counterfactual evaluation metrics. The numbers in the table represent mean ± s.d. BV refers to validity before filtering generated molecules, while SA represents synthetic accessibility, with lower scores indicating easier synthesis.}
    \centering
    \label{tab:counterfactual_explanations}
    \resizebox{0.875\textwidth}{!}{
    \begin{tabular}{l p{32mm}>{\centering\arraybackslash}p{20mm}>{\centering\arraybackslash}p{20mm}>{\centering\arraybackslash}p{20mm}>{\centering\arraybackslash}p{20mm}>
    {\centering\arraybackslash}p{20mm}}
    \toprule
    Dataset & Method & BV~$\uparrow$ & Validity~$\uparrow$ & Similarity~$\uparrow$ & Diversity~$\uparrow$ & SA~$\downarrow$  \\
    \midrule
    \multirow{5}{*}{\rotatebox{0}{\parbox{11mm}{CYP\\3A4}}} 
      & GNNExplainer       & $0.16 \pm 0.04$      & $0.16 \pm 0.04$       & $0.33 \pm 0.00$       & $0.00 \pm 0.00$      & $3.67 \pm 0.04$ \\
      & NN            & $1.00 \pm 0.00$      & $1.00 \pm 0.00$       & $0.37 \pm 0.00$       & $0.66 \pm 0.00$      & $2.53 \pm 0.01$ \\
      \cmidrule(lr){2-7}
      & MMACE              & $0.27 \pm 0.00$      & $\mathbf{1.00 \pm 0.00}$ & $\mathbf{0.60 \pm 0.01}$ & $\mathbf{0.52 \pm 0.01}$       & $3.34 \pm 0.02$ \\
      & CM (CReM)         & $0.21 \pm 0.01$      & $\mathbf{1.00 \pm 0.00}$ & $0.52 \pm 0.04$       & $0.46 \pm 0.04$ &   $\mathbf{2.71 \pm 0.08}$   \\
      & CM (DiffLinker)        & $\mathbf{0.41 \pm 0.02}$      & $\mathbf{1.00 \pm 0.00}$ & $0.41 \pm 0.05$       & $0.41 \pm 0.01$     & $3.91 \pm 0.18$ \\
    \midrule
    \multirow{5}{*}{\rotatebox{0}{\parbox{11mm}{CYP\\2D6}}}
       & GNNExplainer       & $0.10 \pm 0.10$      & $0.10 \pm 0.10$       & $0.35 \pm 0.05$       & $0.00 \pm 0.00$      & $3.62 \pm 0.19$ \\
      & NN            & $1.00 \pm 0.00$      & $1.00 \pm 0.00$       & $0.32 \pm 0.00$       & $0.68 \pm 0.00$      & $2.53 \pm 0.02$ \\
      \cmidrule(lr){2-7}
      & MMACE              & $\mathbf{0.18 \pm 0.00}$      & $\mathbf{1.00 \pm 0.00}$ & $\mathbf{0.52 \pm 0.00}$ & $\mathbf{0.57 \pm 0.00}$       & $3.45 \pm 0.03$ \\
      & CM (CReM)         & $0.08 \pm 0.00$      & $\mathbf{1.00 \pm 0.00}$ & $0.44 \pm 0.01$       & $0.49 \pm 0.01$ &   $\mathbf{2.79 \pm 0.04}$   \\
      & CM (DiffLinker)        & $0.15 \pm 0.02$      & $\mathbf{1.00 \pm 0.00}$ & $0.46 \pm 0.01$       & $0.38 \pm 0.02$     & $3.91 \pm 0.16$ \\
    \midrule
    \multirow{5}{*}{\rotatebox{0}{\parbox{11mm}{hERG}}}
      & GNNExplainer       & $0.21 \pm 0.06$      & $0.21 \pm 0.06$       & $0.31 \pm 0.08$       & $0.00 \pm 0.00$      & $4.20 \pm 0.05$ \\
      & NN            & $1.00 \pm 0.00$      & $1.00 \pm 0.00$       & $0.30 \pm 0.02$       & $0.67 \pm 0.02$      & $3.14 \pm 0.09$ \\
      \cmidrule(lr){2-7}
      & MMACE              & $0.39 \pm 0.02$      & $\mathbf{1.00 \pm 0.00}$ & $\mathbf{0.65 \pm 0.01}$ & $\mathbf{0.48 \pm 0.01}$       & $3.69 \pm 0.02$ \\
      & CM (CReM)         & $0.30 \pm 0.07$      & $\mathbf{1.00 \pm 0.00}$ & $0.51 \pm 0.03$       & $0.47 \pm 0.04$ &   $\mathbf{3.10 \pm 0.09}$   \\
      & CM (DiffLinker)         & $\mathbf{0.61 \pm 0.07}$      & $\mathbf{1.00 \pm 0.00}$ & $0.41 \pm 0.01$       & $\mathbf{0.48 \pm 0.01}$     & $3.63 \pm 0.14$ \\
    \bottomrule
    \end{tabular}
    } 
\end{table*}

\begin{figure}[tb]
    \centering
    \includegraphics[width=1\linewidth]{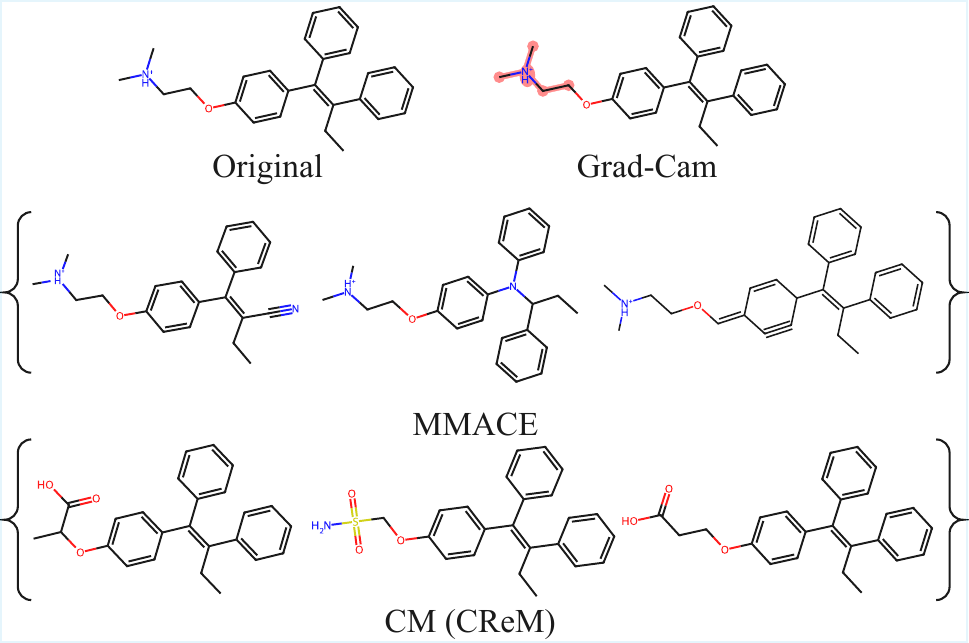}
    \caption{Comparison of counterfactual examples generated via MMACE and CM (CReM). The initial model prediction indicated that the original molecule is a hERG blocker.}
    \label{fig:counterfactuals_examples}
\end{figure}

Figure~\ref{fig:counterfactuals_examples} displays counterfactuals generated by MMACE and CM (CReM) for a hERG inhibitor. The top 20\% most influential atoms identified by Grad-CAM point to a protonated tertiary amine, a common structural feature of hERG blockers~\citep{garrido2020herg}. MMACE often produces structurally complex, hard-to-interpret counterfactuals, such as a triple bond in a ring. In contrast, CM (CReM) creates chemically plausible, targeted modifications. For example, CReM replaces a positively charged tertiary amine with neutral groups, potentially reducing hERG inhibition risk by limiting interactions with aromatic amino acids. Additional examples can be found in the Appendix C. 

\subsection{Improved benchmark of XAI methods}

In the widely used fidelity metric, the predictions are made for the original graph and the graph with masked important atoms. Then, the difference in accuracy can be measured to assess the robustness of the explanation method. However, masking atom features with zeros creates out-of-distribution molecular graphs, which do not accurately reflect the importance of these atoms. The messages in such graphs can still be passed through masked nodes in graph neural networks. It is thus desirable to be able to evaluate explainers with alternative masking methods, and our generative counterfactual approach addresses this necessity. CM can be used to measure the quality of the factual explanation.

\subsubsection{Datasets and models.}

We used five publicly available datasets from the TDC. Three datasets were for binary classification tasks: CYP2D6, CYP3A4, and hERG; two were for regression tasks: predicting Lipophilicity and Solubility in water. 
As a preprocessing step, molecules were stripped of ions and molecules with fewer than five heavy atoms were removed.
We evaluated six distinct graph neural network architectures and regularizations: three variants of the Graph Isomorphism Network (GIN): large GIN (%
2,648,065
parameters trained at 30\% dropout); medium GIN (%
170,497
parameters at 15\% dropout), and small GIN (%
11,905
parameters, no dropout), one GIN model with additional edge attributes (%
106,189
parameters at 10\% dropout), one GIN model with residual connections (%
56,577
parameters at 10\% dropout), and one Graph Attention Network model~\citep{velivckovic2018graph} (3 attention heads, 176,385 parameters at 10\% dropout). A detailed definition of the model architectures is presented in the Appendix A.
For each dataset and model architecture, we performed three independent trainings.
In each training, the data was randomly split into training (75\%), validation (10\%), and test (15\%) sets.
Models were trained using the Adam optimizer with an initial learning rate of $3\times 10^{-4}$
and a batch size of 64 molecules.
The learning rate was reduced by a factor of 0.85 every 10 epochs.
Binary cross-entropy and smooth L1 were used as loss functions for classification and regression tasks, respectively.
Training was carried out for a maximum of 300 epochs, with an early stopping condition that terminated training if the validation loss did not improve for 50 consecutive epochs.
In this way, for each of the five chemical datasets, we obtained 18 trained models and corresponding test sets.

\begin{table*}[tb]
  \caption{Consistency of masking methods and explanation methods.
  The numbers in the table represent mean $\pm$ s.d. of consistency scores, which are the percentage of times that CM of influential atoms (as identified by an explainer) caused the model's prediction to change in the expected direction. A higher score means the combination of the explainer and masking method is more effective and reliable. A score of 50\% is the random baseline, indicating no better than chance.}
  \label{tab:expl-cf-benchmark}
  \centering
\resizebox{0.75\textwidth}{!}{
\begin{tabular}{llp{17mm}p{17mm}p{17mm}p{17mm}p{17mm}}
\toprule
   \multirow{2}{*}{\parbox{20mm}{Masking method}}
 & \multirow{2}{*}{\parbox{20mm}{Explanation method}}
 & \multicolumn{5}{c}{Dataset} \\
  \cmidrule{3-7}
 & & CYP 2D6 & CYP 3A4 & Lipophilicity & Solubility & hERG   \\
\midrule
\multirow[c]{5}{*}{Feature zeroing} & Grad-CAM & \tikz[baseline]{
  \fill[gray!29] (0, -0.09) rectangle (1.5085289711041558, 0.34);
  \node[anchor=west, font=\normalsize] at (0, 0.1) {84{\footnotesize\textcolor{darkgray}{$\,\pm\,$13}}\%};
}
 & \tikz[baseline]{
  \fill[gray!29] (0, -0.09) rectangle (1.4273411752339704, 0.34);
  \node[anchor=west, font=\normalsize] at (0, 0.1) {79{\footnotesize\textcolor{darkgray}{$\,\pm\,$23}}\%};
}
 & \tikz[baseline]{
  \fill[gray!29] (0, -0.09) rectangle (1.5094431258570962, 0.34);
  \node[anchor=west, font=\normalsize] at (0, 0.1) {84{\footnotesize\textcolor{darkgray}{$\,\pm\,$17}}\%};
}
 & \tikz[baseline]{
  \fill[gray!29] (0, -0.09) rectangle (1.3637164963436754, 0.34);
  \node[anchor=west, font=\normalsize] at (0, 0.1) {76{\footnotesize\textcolor{darkgray}{$\,\pm\,$23}}\%};
}
 & \tikz[baseline]{
  \fill[gray!29] (0, -0.09) rectangle (1.4225618849120487, 0.34);
  \node[anchor=west, font=\normalsize] at (0, 0.1) {79{\footnotesize\textcolor{darkgray}{$\,\pm\,$21}}\%};
}
 \\
 & Integrated Gradients & \tikz[baseline]{
  \fill[gray!29] (0, -0.09) rectangle (1.3600303249071248, 0.34);
  \node[anchor=west, font=\normalsize] at (0, 0.1) {76{\footnotesize\textcolor{darkgray}{$\,\pm\,$13}}\%};
}
 & \tikz[baseline]{
  \fill[gray!29] (0, -0.09) rectangle (1.5066408003537104, 0.34);
  \node[anchor=west, font=\normalsize] at (0, 0.1) {84{\footnotesize\textcolor{darkgray}{$\,\pm\,$14}}\%};
}
 & \tikz[baseline]{
  \fill[gray!29] (0, -0.09) rectangle (1.4491631457655678, 0.34);
  \node[anchor=west, font=\normalsize] at (0, 0.1) {81{\footnotesize\textcolor{darkgray}{$\,\pm\,$17}}\%};
}
 & \tikz[baseline]{
  \fill[gray!29] (0, -0.09) rectangle (1.3831444576957266, 0.34);
  \node[anchor=west, font=\normalsize] at (0, 0.1) {77{\footnotesize\textcolor{darkgray}{$\,\pm\,$12}}\%};
}
 & \tikz[baseline]{
  \fill[gray!29] (0, -0.09) rectangle (1.2540275394378086, 0.34);
  \node[anchor=west, font=\normalsize] at (0, 0.1) {70{\footnotesize\textcolor{darkgray}{$\,\pm\,$17}}\%};
}
 \\
 & GNNExplainer & \tikz[baseline]{
  \fill[gray!29] (0, -0.09) rectangle (1.1109962445555748, 0.34);
  \node[anchor=west, font=\normalsize] at (0, 0.1) {62{\footnotesize\textcolor{darkgray}{$\,\pm\,$22}}\%};
}
 & \tikz[baseline]{
  \fill[gray!29] (0, -0.09) rectangle (1.0507045233733905, 0.34);
  \node[anchor=west, font=\normalsize] at (0, 0.1) {58{\footnotesize\textcolor{darkgray}{$\,\pm\,$35}}\%};
}
 & \tikz[baseline]{
  \fill[gray!29] (0, -0.09) rectangle (0.9818347058964273, 0.34);
  \node[anchor=west, font=\normalsize] at (0, 0.1) {55{\footnotesize\textcolor{darkgray}{$\,\pm\,$41}}\%};
}
 & \tikz[baseline]{
  \fill[gray!29] (0, -0.09) rectangle (1.0163764344582922, 0.34);
  \node[anchor=west, font=\normalsize] at (0, 0.1) {56{\footnotesize\textcolor{darkgray}{$\,\pm\,$32}}\%};
}
 & \tikz[baseline]{
  \fill[gray!29] (0, -0.09) rectangle (1.011765517517902, 0.34);
  \node[anchor=west, font=\normalsize] at (0, 0.1) {56{\footnotesize\textcolor{darkgray}{$\,\pm\,$38}}\%};
}
 \\
 & Saliency & \tikz[baseline]{
  \fill[gray!29] (0, -0.09) rectangle (1.0064612790388585, 0.34);
  \node[anchor=west, font=\normalsize] at (0, 0.1) {56{\footnotesize\textcolor{darkgray}{$\,\pm\,$19}}\%};
}
 & \tikz[baseline]{
  \fill[gray!29] (0, -0.09) rectangle (1.1367167130055733, 0.34);
  \node[anchor=west, font=\normalsize] at (0, 0.1) {63{\footnotesize\textcolor{darkgray}{$\,\pm\,$31}}\%};
}
 & \tikz[baseline]{
  \fill[gray!29] (0, -0.09) rectangle (1.0850077490772134, 0.34);
  \node[anchor=west, font=\normalsize] at (0, 0.1) {60{\footnotesize\textcolor{darkgray}{$\,\pm\,$30}}\%};
}
 & \tikz[baseline]{
  \fill[gray!29] (0, -0.09) rectangle (0.9438764773201994, 0.34);
  \node[anchor=west, font=\normalsize] at (0, 0.1) {52{\footnotesize\textcolor{darkgray}{$\,\pm\,$25}}\%};
}
 & \tikz[baseline]{
  \fill[gray!29] (0, -0.09) rectangle (0.9597703917652569, 0.34);
  \node[anchor=west, font=\normalsize] at (0, 0.1) {53{\footnotesize\textcolor{darkgray}{$\,\pm\,$31}}\%};
}
 \\
 & Random & \tikz[baseline]{
  \fill[gray!29] (0, -0.09) rectangle (0.9017746442173031, 0.34);
  \node[anchor=west, font=\normalsize] at (0, 0.1) {50{\footnotesize\textcolor{darkgray}{$\,\pm\,$16}}\%};
}
 & \tikz[baseline]{
  \fill[gray!29] (0, -0.09) rectangle (0.898022778828589, 0.34);
  \node[anchor=west, font=\normalsize] at (0, 0.1) {50{\footnotesize\textcolor{darkgray}{$\,\pm\,$26}}\%};
}
 & \tikz[baseline]{
  \fill[gray!29] (0, -0.09) rectangle (0.8975292824802914, 0.34);
  \node[anchor=west, font=\normalsize] at (0, 0.1) {50{\footnotesize\textcolor{darkgray}{$\,\pm\,$27}}\%};
}
 & \tikz[baseline]{
  \fill[gray!29] (0, -0.09) rectangle (0.9026025273073147, 0.34);
  \node[anchor=west, font=\normalsize] at (0, 0.1) {50{\footnotesize\textcolor{darkgray}{$\,\pm\,$19}}\%};
}
 & \tikz[baseline]{
  \fill[gray!29] (0, -0.09) rectangle (0.902919767609582, 0.34);
  \node[anchor=west, font=\normalsize] at (0, 0.1) {50{\footnotesize\textcolor{darkgray}{$\,\pm\,$30}}\%};
}
 \\
\cmidrule(lr){1-7}
\multirow[c]{5}{*}{CM (DiffLinker)} & Grad-CAM & \tikz[baseline]{
  \fill[gray!29] (0, -0.09) rectangle (1.006388024066225, 0.34);
  \node[anchor=west, font=\normalsize] at (0, 0.1) {56{\footnotesize\textcolor{darkgray}{$\,\pm\,$28}}\%};
}
 & \tikz[baseline]{
  \fill[gray!29] (0, -0.09) rectangle (1.053923530368644, 0.34);
  \node[anchor=west, font=\normalsize] at (0, 0.1) {59{\footnotesize\textcolor{darkgray}{$\,\pm\,$21}}\%};
}
 & \tikz[baseline]{
  \fill[gray!29] (0, -0.09) rectangle (1.0047772817213343, 0.34);
  \node[anchor=west, font=\normalsize] at (0, 0.1) {56{\footnotesize\textcolor{darkgray}{$\,\pm\,$38}}\%};
}
 & \tikz[baseline]{
  \fill[gray!29] (0, -0.09) rectangle (0.9907708751757626, 0.34);
  \node[anchor=west, font=\normalsize] at (0, 0.1) {55{\footnotesize\textcolor{darkgray}{$\,\pm\,$36}}\%};
}
 & \tikz[baseline]{
  \fill[gray!29] (0, -0.09) rectangle (0.9679522681064348, 0.34);
  \node[anchor=west, font=\normalsize] at (0, 0.1) {54{\footnotesize\textcolor{darkgray}{$\,\pm\,$40}}\%};
}
 \\
 & Integrated Gradients & \tikz[baseline]{
  \fill[gray!29] (0, -0.09) rectangle (0.9629962213691938, 0.34);
  \node[anchor=west, font=\normalsize] at (0, 0.1) {53{\footnotesize\textcolor{darkgray}{$\,\pm\,$30}}\%};
}
 & \tikz[baseline]{
  \fill[gray!29] (0, -0.09) rectangle (0.9658806106535283, 0.34);
  \node[anchor=west, font=\normalsize] at (0, 0.1) {54{\footnotesize\textcolor{darkgray}{$\,\pm\,$24}}\%};
}
 & \tikz[baseline]{
  \fill[gray!29] (0, -0.09) rectangle (0.9753515382805514, 0.34);
  \node[anchor=west, font=\normalsize] at (0, 0.1) {54{\footnotesize\textcolor{darkgray}{$\,\pm\,$39}}\%};
}
 & \tikz[baseline]{
  \fill[gray!29] (0, -0.09) rectangle (0.9676579995185727, 0.34);
  \node[anchor=west, font=\normalsize] at (0, 0.1) {54{\footnotesize\textcolor{darkgray}{$\,\pm\,$40}}\%};
}
 & \tikz[baseline]{
  \fill[gray!29] (0, -0.09) rectangle (0.9523883102612104, 0.34);
  \node[anchor=west, font=\normalsize] at (0, 0.1) {53{\footnotesize\textcolor{darkgray}{$\,\pm\,$37}}\%};
}
 \\
 & GNNExplainer & \tikz[baseline]{
  \fill[gray!29] (0, -0.09) rectangle (0.9412082603230529, 0.34);
  \node[anchor=west, font=\normalsize] at (0, 0.1) {52{\footnotesize\textcolor{darkgray}{$\,\pm\,$28}}\%};
}
 & \tikz[baseline]{
  \fill[gray!29] (0, -0.09) rectangle (0.937547008647845, 0.34);
  \node[anchor=west, font=\normalsize] at (0, 0.1) {52{\footnotesize\textcolor{darkgray}{$\,\pm\,$24}}\%};
}
 & \tikz[baseline]{
  \fill[gray!29] (0, -0.09) rectangle (0.9116302378800891, 0.34);
  \node[anchor=west, font=\normalsize] at (0, 0.1) {51{\footnotesize\textcolor{darkgray}{$\,\pm\,$36}}\%};
}
 & \tikz[baseline]{
  \fill[gray!29] (0, -0.09) rectangle (0.942169989294756, 0.34);
  \node[anchor=west, font=\normalsize] at (0, 0.1) {52{\footnotesize\textcolor{darkgray}{$\,\pm\,$38}}\%};
}
 & \tikz[baseline]{
  \fill[gray!29] (0, -0.09) rectangle (0.9224460988667232, 0.34);
  \node[anchor=west, font=\normalsize] at (0, 0.1) {51{\footnotesize\textcolor{darkgray}{$\,\pm\,$41}}\%};
}
 \\
 & Saliency & \tikz[baseline]{
  \fill[gray!29] (0, -0.09) rectangle (0.9046260091637749, 0.34);
  \node[anchor=west, font=\normalsize] at (0, 0.1) {50{\footnotesize\textcolor{darkgray}{$\,\pm\,$30}}\%};
}
 & \tikz[baseline]{
  \fill[gray!29] (0, -0.09) rectangle (0.8904384375793767, 0.34);
  \node[anchor=west, font=\normalsize] at (0, 0.1) {49{\footnotesize\textcolor{darkgray}{$\,\pm\,$23}}\%};
}
 & \tikz[baseline]{
  \fill[gray!29] (0, -0.09) rectangle (0.9381088281390481, 0.34);
  \node[anchor=west, font=\normalsize] at (0, 0.1) {52{\footnotesize\textcolor{darkgray}{$\,\pm\,$38}}\%};
}
 & \tikz[baseline]{
  \fill[gray!29] (0, -0.09) rectangle (0.9096036749992934, 0.34);
  \node[anchor=west, font=\normalsize] at (0, 0.1) {51{\footnotesize\textcolor{darkgray}{$\,\pm\,$39}}\%};
}
 & \tikz[baseline]{
  \fill[gray!29] (0, -0.09) rectangle (0.8973995101434618, 0.34);
  \node[anchor=west, font=\normalsize] at (0, 0.1) {50{\footnotesize\textcolor{darkgray}{$\,\pm\,$39}}\%};
}
 \\
 & Random & \tikz[baseline]{
  \fill[gray!29] (0, -0.09) rectangle (0.8961902325613798, 0.34);
  \node[anchor=west, font=\normalsize] at (0, 0.1) {50{\footnotesize\textcolor{darkgray}{$\,\pm\,$28}}\%};
}
 & \tikz[baseline]{
  \fill[gray!29] (0, -0.09) rectangle (0.9020132497806683, 0.34);
  \node[anchor=west, font=\normalsize] at (0, 0.1) {50{\footnotesize\textcolor{darkgray}{$\,\pm\,$24}}\%};
}
 & \tikz[baseline]{
  \fill[gray!29] (0, -0.09) rectangle (0.8927474074673386, 0.34);
  \node[anchor=west, font=\normalsize] at (0, 0.1) {50{\footnotesize\textcolor{darkgray}{$\,\pm\,$37}}\%};
}
 & \tikz[baseline]{
  \fill[gray!29] (0, -0.09) rectangle (0.8997674284681664, 0.34);
  \node[anchor=west, font=\normalsize] at (0, 0.1) {50{\footnotesize\textcolor{darkgray}{$\,\pm\,$39}}\%};
}
 & \tikz[baseline]{
  \fill[gray!29] (0, -0.09) rectangle (0.9073902834263343, 0.34);
  \node[anchor=west, font=\normalsize] at (0, 0.1) {50{\footnotesize\textcolor{darkgray}{$\,\pm\,$39}}\%};
}
 \\
\cmidrule(lr){1-7}
\multirow[c]{5}{*}{CM (CReM)} & Grad-CAM & \tikz[baseline]{
  \fill[gray!29] (0, -0.09) rectangle (1.2690929603470738, 0.34);
  \node[anchor=west, font=\normalsize] at (0, 0.1) {71{\footnotesize\textcolor{darkgray}{$\,\pm\,$10}}\%};
}
 & \tikz[baseline]{
  \fill[gray!29] (0, -0.09) rectangle (1.294179879783577, 0.34);
  \node[anchor=west, font=\normalsize] at (0, 0.1) {72{\footnotesize\textcolor{darkgray}{$\,\pm\,$13}}\%};
}
 & \tikz[baseline]{
  \fill[gray!29] (0, -0.09) rectangle (1.3708926452499433, 0.34);
  \node[anchor=west, font=\normalsize] at (0, 0.1) {76{\footnotesize\textcolor{darkgray}{$\,\pm\,$14}}\%};
}
 & \tikz[baseline]{
  \fill[gray!29] (0, -0.09) rectangle (1.1364188461956253, 0.34);
  \node[anchor=west, font=\normalsize] at (0, 0.1) {63{\footnotesize\textcolor{darkgray}{$\,\pm\,$21}}\%};
}
 & \tikz[baseline]{
  \fill[gray!29] (0, -0.09) rectangle (1.2949744514794947, 0.34);
  \node[anchor=west, font=\normalsize] at (0, 0.1) {72{\footnotesize\textcolor{darkgray}{$\,\pm\,$16}}\%};
}
 \\
 & Integrated Gradients & \tikz[baseline]{
  \fill[gray!29] (0, -0.09) rectangle (0.9339866997477002, 0.34);
  \node[anchor=west, font=\normalsize] at (0, 0.1) {52{\footnotesize\textcolor{darkgray}{$\,\pm\,$13}}\%};
}
 & \tikz[baseline]{
  \fill[gray!29] (0, -0.09) rectangle (1.0067545586039974, 0.34);
  \node[anchor=west, font=\normalsize] at (0, 0.1) {56{\footnotesize\textcolor{darkgray}{$\,\pm\,$17}}\%};
}
 & \tikz[baseline]{
  \fill[gray!29] (0, -0.09) rectangle (1.0965889711514654, 0.34);
  \node[anchor=west, font=\normalsize] at (0, 0.1) {61{\footnotesize\textcolor{darkgray}{$\,\pm\,$12}}\%};
}
 & \tikz[baseline]{
  \fill[gray!29] (0, -0.09) rectangle (1.2243346279720153, 0.34);
  \node[anchor=west, font=\normalsize] at (0, 0.1) {68{\footnotesize\textcolor{darkgray}{$\,\pm\,$14}}\%};
}
 & \tikz[baseline]{
  \fill[gray!29] (0, -0.09) rectangle (0.9779462688671964, 0.34);
  \node[anchor=west, font=\normalsize] at (0, 0.1) {54{\footnotesize\textcolor{darkgray}{$\,\pm\,$21}}\%};
}
 \\
 & GNNExplainer & \tikz[baseline]{
  \fill[gray!29] (0, -0.09) rectangle (1.072892452076564, 0.34);
  \node[anchor=west, font=\normalsize] at (0, 0.1) {60{\footnotesize\textcolor{darkgray}{$\,\pm\,$14}}\%};
}
 & \tikz[baseline]{
  \fill[gray!29] (0, -0.09) rectangle (1.0668839778989048, 0.34);
  \node[anchor=west, font=\normalsize] at (0, 0.1) {59{\footnotesize\textcolor{darkgray}{$\,\pm\,$21}}\%};
}
 & \tikz[baseline]{
  \fill[gray!29] (0, -0.09) rectangle (1.0265835498828229, 0.34);
  \node[anchor=west, font=\normalsize] at (0, 0.1) {57{\footnotesize\textcolor{darkgray}{$\,\pm\,$28}}\%};
}
 & \tikz[baseline]{
  \fill[gray!29] (0, -0.09) rectangle (0.9301686873281888, 0.34);
  \node[anchor=west, font=\normalsize] at (0, 0.1) {52{\footnotesize\textcolor{darkgray}{$\,\pm\,$30}}\%};
}
 & \tikz[baseline]{
  \fill[gray!29] (0, -0.09) rectangle (1.0510733064611086, 0.34);
  \node[anchor=west, font=\normalsize] at (0, 0.1) {58{\footnotesize\textcolor{darkgray}{$\,\pm\,$21}}\%};
}
 \\
 & Saliency & \tikz[baseline]{
  \fill[gray!29] (0, -0.09) rectangle (0.9202935664031111, 0.34);
  \node[anchor=west, font=\normalsize] at (0, 0.1) {51{\footnotesize\textcolor{darkgray}{$\,\pm\,$12}}\%};
}
 & \tikz[baseline]{
  \fill[gray!29] (0, -0.09) rectangle (0.9451767365241791, 0.34);
  \node[anchor=west, font=\normalsize] at (0, 0.1) {53{\footnotesize\textcolor{darkgray}{$\,\pm\,$19}}\%};
}
 & \tikz[baseline]{
  \fill[gray!29] (0, -0.09) rectangle (0.9585517201952691, 0.34);
  \node[anchor=west, font=\normalsize] at (0, 0.1) {53{\footnotesize\textcolor{darkgray}{$\,\pm\,$17}}\%};
}
 & \tikz[baseline]{
  \fill[gray!29] (0, -0.09) rectangle (0.8780707392839469, 0.34);
  \node[anchor=west, font=\normalsize] at (0, 0.1) {49{\footnotesize\textcolor{darkgray}{$\,\pm\,$24}}\%};
}
 & \tikz[baseline]{
  \fill[gray!29] (0, -0.09) rectangle (0.9192790532772193, 0.34);
  \node[anchor=west, font=\normalsize] at (0, 0.1) {51{\footnotesize\textcolor{darkgray}{$\,\pm\,$18}}\%};
}
 \\
 & Random & \tikz[baseline]{
  \fill[gray!29] (0, -0.09) rectangle (0.8952600354760738, 0.34);
  \node[anchor=west, font=\normalsize] at (0, 0.1) {50{\footnotesize\textcolor{darkgray}{$\,\pm\,$13}}\%};
}
 & \tikz[baseline]{
  \fill[gray!29] (0, -0.09) rectangle (0.8932381158635349, 0.34);
  \node[anchor=west, font=\normalsize] at (0, 0.1) {50{\footnotesize\textcolor{darkgray}{$\,\pm\,$18}}\%};
}
 & \tikz[baseline]{
  \fill[gray!29] (0, -0.09) rectangle (0.903649641584372, 0.34);
  \node[anchor=west, font=\normalsize] at (0, 0.1) {50{\footnotesize\textcolor{darkgray}{$\,\pm\,$22}}\%};
}
 & \tikz[baseline]{
  \fill[gray!29] (0, -0.09) rectangle (0.8864673278210982, 0.34);
  \node[anchor=west, font=\normalsize] at (0, 0.1) {49{\footnotesize\textcolor{darkgray}{$\,\pm\,$28}}\%};
}
 & \tikz[baseline]{
  \fill[gray!29] (0, -0.09) rectangle (0.9041363694005975, 0.34);
  \node[anchor=west, font=\normalsize] at (0, 0.1) {50{\footnotesize\textcolor{darkgray}{$\,\pm\,$22}}\%};
}
 \\
\bottomrule
\end{tabular}
}
\end{table*}

\subsubsection{Experimental setup.}

For each combination of masking techniques and datasets, we used five explainability methods to identify the atoms most responsible for a specific prediction: Grad-CAM, Integrated Gradients, GNNExplainer, saliency map, and, as a baseline, random assignment. For each molecule, we determined two distinct sets of atoms: the top 10\% that most strongly increased the predicted value and, separately, the top 10\% that most strongly decreased it. GNNExplainer was configured to run 100 iterations and mask node attributes, with the influence direction determined by an auxiliary Grad-CAM assessment. The influential atoms identified were then masked using one of three techniques: (1) CM (CReM), (2) CM (DiffLinker), or (3) simple atom feature ablation (setting features to zero).

To ensure the validity of our analysis, we first filtered the generated molecules. When the new, replacement atoms were indicated by an explainer as contributing to the prediction in the same direction but to an even greater extent than the original atoms, the generated molecule was discarded.
In this way, the remaining molecules were expected to have their most influential atoms masked counterfactually,
meaning that the masked region's original influence has been attenuated or reversed.
This selection procedure was applied to molecules generated by all three masking techniques.
We then assessed whether masking influential atoms led to the expected change in the model's overall prediction.
We calculated a consistency score, defined as the proportion of counterfactually masked molecules where the property prediction changed in the direction anticipated by the explainer (e.g., the prediction decreased after masking atoms identified as increasing the value).
This consistency score was calculated for 3 independent data splits $\times$ 6 model architectures $\times$ 2 directions of the change in a predicted molecular property.

\subsubsection{Benchmark results.}

Table~\ref{tab:expl-cf-benchmark} reports the aggregate statistics for consistency scores, summarizing the results from every combination of the 5 (test) datasets, 5 explanation methods, and 3 masking techniques. Statistics are computed for 300 randomly selected  molecules from each test dataset. Detailed results for separate models are provided in the supplementary tables in the Appendix D.

Across all five datasets, the combination of the Grad-CAM explainer with the standard masking by feature zeroing consistently yields the highest average consistency scores (ranging from 76\% to 84\%). However, it should be remembered that this method masks atom features but leaves the graph structure intact, potentially leaking information and inflating consistency scores. CReM, acting through molecular replacement under more stringent chemical requirements, still achieves strong consistency scores, particularly with Grad-CAM (ranging from 63\% to 76\%). These results are well above the 50\% random baseline (and above the results for DiffLinker), demonstrating that CReM effectively creates meaningful counterfactuals that validate the explainer's output. A qualitative example of compound solubility analysis and explanation method evaluation using CM is provided in Appendix E.

\subsection{Limitations}

Although CM brings chemical realism to XAI model evaluation, it has important limitations that should be addressed in future research. One significant limitation is the validity of molecules generated by current models conditioned on molecular fragment context. CReM struggles to fill in fragments with more than two attachment points. Conversely, DiffLinker can even complete partial rings but often fails to produce chemically feasible molecules. Because sampling multiple compounds from generative models is necessary, CM also demands more computational resources for model evaluation.

\section{Conclusions}

In this paper, we introduced CM, a method that combines factual explanation techniques for molecular graphs with structure-conditioned generative models to generate a set of counterfactual examples. This approach improves the evaluation of XAI methods for molecular property prediction by creating molecules with fragment replacements that serve as better references than masked molecules with artificially removed nodes or features. Additionally, we demonstrate how CM can be used as a counterfactual explanation technique that performs precise, local replacements of fragments essential for the model's prediction. We hope that our method will advance the development of more effective XAI methods and accelerate molecular design workflows.

\section{Acknowledgments}

This study was funded by the "Interpretable and Interactive Multimodal Retrieval in Drug Discovery" project. The "Interpretable and Interactive Multimodal Retrieval in Drug Discovery” project (FENG.02.02-IP.05-0040/23) is carried out within the First Team programme of the Foundation for Polish Science co-financed by the European Union under the European Funds for Smart Economy 2021-2027 (FENG). We gratefully acknowledge Polish high-performance computing infrastructure PLGrid (HPC Center: ACK Cyfronet AGH) for providing computer facilities and support within computational grant no. PLG/2025/018272.

\bibliography{references}

\renewcommand{\thetable}{A\arabic{table}}
\renewcommand{\thefigure}{A\arabic{figure}}
\input{appendix.tex}

\end{document}

%% file: appendix.tex
\appendix

\setcounter{figure}{0}
\setcounter{table}{0}

\section{A. Training details}


 \subsubsection{Python environment.} All experiments were carried out using Python 3.11.5. The key libraries are listed in Table~\ref{tab:libraries}. Other requirements are included in the code repository.

\begin{table}[h]
\centering
\caption{\textbf{Key libraries used in the experiments.}}
\label{tab:libraries}
\begin{tabular}{ll}
\toprule
\textbf{Library}                 & \textbf{Version} \\
\midrule
{PyTorch}                          & 2.5.1+cu124      \\ 
{PyTorch Lightning}                & 2.5.1            \\ 
{TorchGeometric}                   & 2.6.1            \\ 
{RDKit}                            & 2023.9.6         \\ 
{OpenBabel}                        & 3.1.1.22         \\ 
{ExMol}                            & 3.3.0            \\ 
{CReM}                             & 0.2.14           \\ 
{PyTDC}                            & 0.4.17           \\ 
{Scikit-learn}                     & 1.5.2            \\ 
{Pandas}                           & 2.2.3            \\ 
{Numpy}                            & 1.26.4           \\ 
\bottomrule
\end{tabular}
\end{table}
 
 \subsubsection{Computational resources.}
 Experiments were conducted on a single NVIDIA Grace 72-core CPU @ 3.1 GHz and a single NVIDIA GH200 GPU with 96 GB of memory, using up to 6.0 GB of system RAM.

\subsubsection{Model architectures.}
Details on the graph node features used in the models, model architectures, and training configurations are provided in Tables~\ref{tab:features}, \ref{tab:model_description}, and \ref{tab:training_configurations}.

\begin{table*} [h]
\centering
\caption{\textbf{Graph node features used in models.}}
\label{tab:features}
\begin{tabular}{@{}lll@{}}
\toprule
\textbf{Feature Type} & \textbf{Description} & \textbf{Encoding Type} \\
\midrule
Atom symbol          & Chemical element (C, O, N, H, S, Cl, F, Other) & One-hot \\
Hydrogen count       & Number of hydrogens attached                   & One-hot \\
Is in ring           & Belongs to any ring structure                  & Binary  \\
Is aromatic          & Aromaticity indicator                          & Binary  \\
Is in 5-member ring  & Member of a 5-atom ring                        & Binary  \\
Is in 6-member ring  & Member of a 6-atom ring                        & Binary  \\
Number of neighbors  & Atom degree (number of bonded atoms)           & One-hot \\
\bottomrule
\end{tabular}
\end{table*}

\begin{table*}[t]
\centering
\caption{\textbf{Model architectures.} }
\label{tab:model_description}
\begin{tabular}{l l c c c c c c c}
\toprule
\textbf{Experiment} & \textbf{Dataset} & \textbf{Model} & \textbf{Task type} & \textbf{Depth} & \textbf{Hidden} & \textbf{Activation} & \textbf{Pooling} & \textbf{Dropout} \\
\midrule
{Pairs} & Solubility & GIN & Regression & 3 & 512 & ReLU & Global Mean & 0.3 \\
\midrule
\multirow{3}{*}{\shortstack{Counter-\\factuals}} 
 & CYP3A4 & GIN & Classification & 3 & 512 & ReLU & Global Mean & 0.3 \\
 & CYP2D6 & GIN & Classification & 3 & 512 & ReLU & Global Mean & 0.3 \\
 & hERG   & GIN & Classification & 3 & 512 & ReLU & Global Mean & 0.3 \\
 \midrule
 \multirow{3}{*}{\shortstack{Explainers}} 
 & \multirow{5}{*}{%
 \parbox{21mm}{\phantom{.}\\[0.25mm]\{CYP3A4, \phantom{\}}CYP2D6, \phantom{\}}hERG, \phantom{\}}Solubility, \phantom{\}}Lipophilicity\}}
 } & LargeGIN  & \multirow{5}{*}{\parbox{22mm}{\phantom{.}\\[0.25mm]\{Classification, \phantom{\{}Classification, \phantom{\{}Classification, \phantom{\{}Regression, \phantom{\{}Regression\}}} & 3 & 512 & ReLU & Global Mean & 0.3 \\
 & & MediumGIN      & & 3 & 128 & ReLU & Global Mean & 0.1 \\
 & & SmallGIN       & & 3 &  32 & ReLU & Global Mean & 0\phantom{.0} \\
 & & EdgeGIN        & & 3 & 128 & ReLU & Global Add & 0.1 \\
 & & ResidualGIN    & & 3 &  64 & ReLU & Global Add & 0.1 \\
 & & GAT (heads: 3) & & 3 &  64 & ReLU & Global Mean & 0.1 \\
\bottomrule
\end{tabular}
\end{table*}

\begin{table*}[t]
\centering
\caption{\textbf{Model training configurations.}}
\label{tab:training_configurations}
\begin{tabular}{l l c c c c c c c}
\toprule
\textbf{Experiment} & \textbf{Dataset} & \textbf{Model} & \textbf{Batch} & \parbox{16mm}{\textbf{\begin{center}Learning Rate\end{center}}} & \textbf{Epochs} & \textbf{Optimizer} & \textbf{Loss} & \parbox{15mm}{\textbf{\begin{center}Early stopping patience\end{center}}} \\
\midrule
Pairs & Solubility & GIN & 64 & 0.001 & 300 & Adam & MSE & -- \\
\midrule
\multirow{3}{*}{\shortstack{Counter-\\factuals}} 
& CYP3A4 & GIN & 16 & 0.001 & 300 & Adam & BCE & 20 \\
& CYP2D6 & GIN & 16 & 0.001 & 300 & Adam & BCE & 20 \\
& hERG   & GIN & 16 & 0.001 & 300 & Adam & BCE & 20 \\
\midrule
 \shortstack{Explainers}
 & \parbox{21mm}{\{CYP3A4, \phantom{\{}CYP2D6, \phantom{\}}hERG, \phantom{\}}Solubility, \phantom{\}}Lipophilicity\}}
 & \parbox{21mm}{\{LargeGIN, \phantom{\{}MediumGIN, \phantom{\{}SmallGIN, \phantom{\{}EdgeGIN, \phantom{\{}ResidualGIN, \phantom{\{}GAT\}}
 & 64 & 0.0003 & 300 & Adam & BCE/Smooth L1 & 50\\
\bottomrule
\end{tabular}
\end{table*}


\section{B. Examples of molecules and distributions of molecular embeddings from the Common Substructure Pair Dataset}

\subsubsection{Examples of molecules.}
\begin{figure*}[t]
  \centering
  \includegraphics[width=1\linewidth]{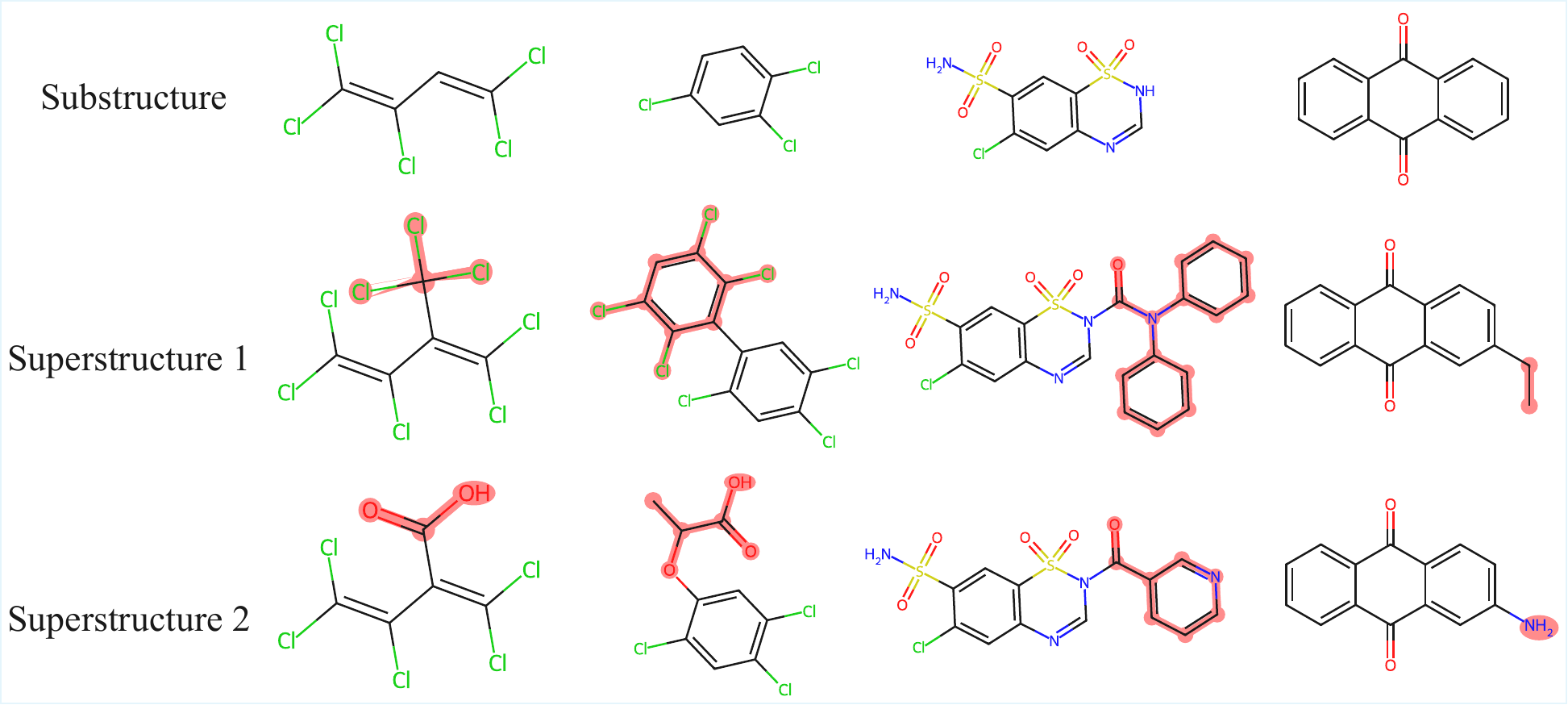}
  \caption{\textbf{Examples of four pairs of molecules with one anchor between the uncommon parts and the common substructure from the Common Substructure Pair Dataset.} The first row shows the shared substructure, while the second and third rows display the superstructures, with the differing fragments highlighted in red.}
  \label{fig:examples_one_anchor}
\end{figure*}

\begin{figure*}[t]
  \centering
  \includegraphics[width=1\linewidth]{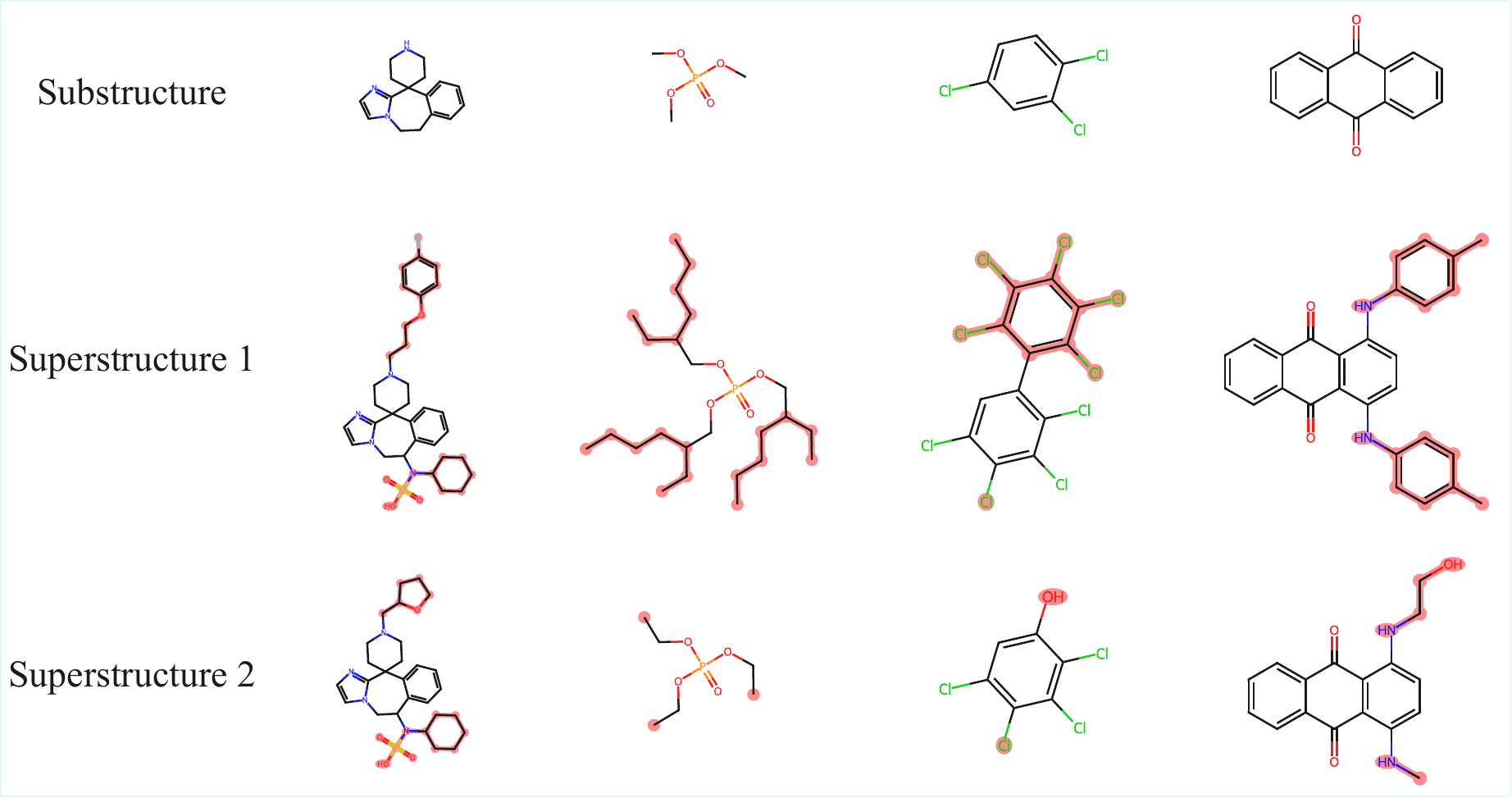}
  \caption{\textbf{Examples of four pairs of molecules with multiple anchors between the uncommon parts and the common substructure from the Common Substructure Pair Dataset.} The first row shows the shared substructure, while the second and third rows display the superstructures, with the differing fragments highlighted in red.}
  \label{fig:examples_multiple_anchors}
\end{figure*}

Figures~\ref{fig:examples_one_anchor} and~\ref{fig:examples_multiple_anchors} illustrate examples of substructures and their corresponding superstructures from the Common Substructure Pair Dataset, categorized by the number of anchors connecting the uncommon parts to the common substructure. In Figure~\ref{fig:examples_one_anchor}, the superstructures are filtered to include only those where the uncommon part is connected to the substructure via a single, identical anchor. In contrast, Figure~\ref{fig:examples_multiple_anchors} shows superstructures in which the uncommon parts are connected through multiple anchors.

Our implementation supports flexible molecule filtering, allowing users to constrain the number or positions of anchors. This enables the generation of diverse subdatasets. On average, each substructure is linked to 52 superstructures, which can be further filtered using user-defined rules.

\subsubsection{Distributions of molecular embeddings.} 
\begin{figure*}
  \centering
  \includegraphics[width=0.9\linewidth]{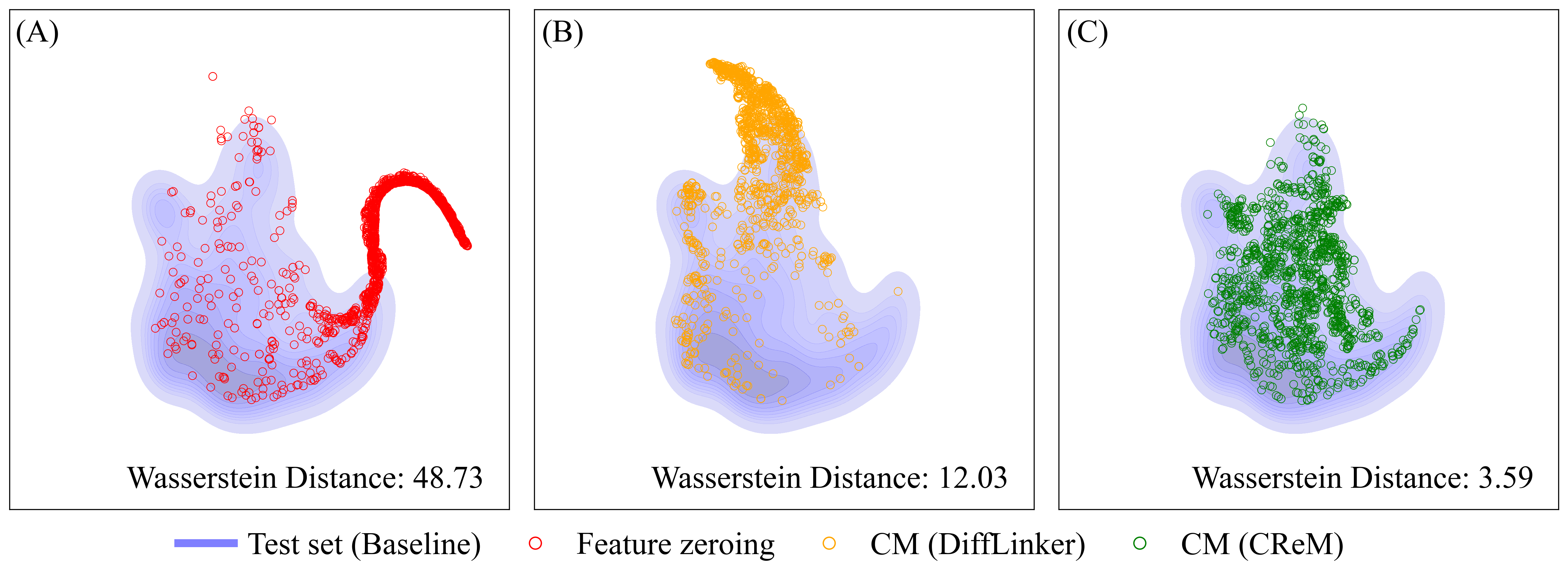}
  \caption{\textbf{t-SNE visualization of molecular embeddings comparing the test set distribution with molecules containing parts masked by Feature zeroing, CM (DiffLinker) and CM(CReM).} Masked molecules come from the multiple anchors part of the dataset (masked fragments have more than one attachment point to the common part). Wasserstein distances quantify the distributional divergence from the test set reference distribution.}
  \label{fig:embedding_comparison_testSet_multiple}
\end{figure*}

\begin{figure*}
  \centering
  \includegraphics[width=0.9\linewidth]{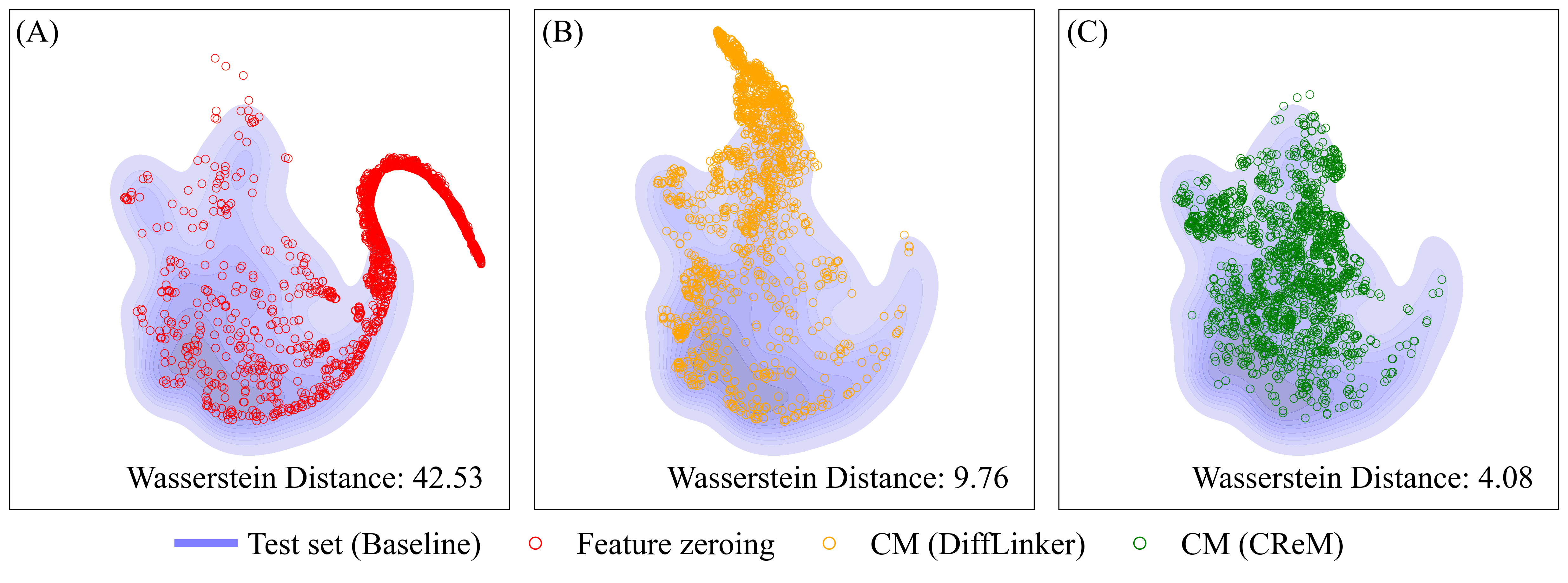}
  \caption{\textbf{t-SNE visualization of molecular embeddings comparing the test set distribution with molecules containing parts masked by Feature zeroing, CM (DiffLinker) and CM(CReM).} Masked molecules come from the "both" part of the dataset (masked fragments can have one or more attachment points to the common part). Wasserstein distances quantify the distributional divergence from the test set reference distribution.}
  \label{fig:embedding_comparison_testSet_both}
\end{figure*}

Figures~\ref{fig:embedding_comparison_testSet_multiple} and~\ref{fig:embedding_comparison_testSet_both} show the distributions of molecular embeddings for molecules with parts masked using Feature zeroing, CM (DiffLinker), and CM (CReM).

\subsubsection{Histograms of prediction differences.} 
\begin{figure*}
\centering
\begin{subfigure}{.5\textwidth}
  \centering
  \includegraphics[width=0.9\linewidth]{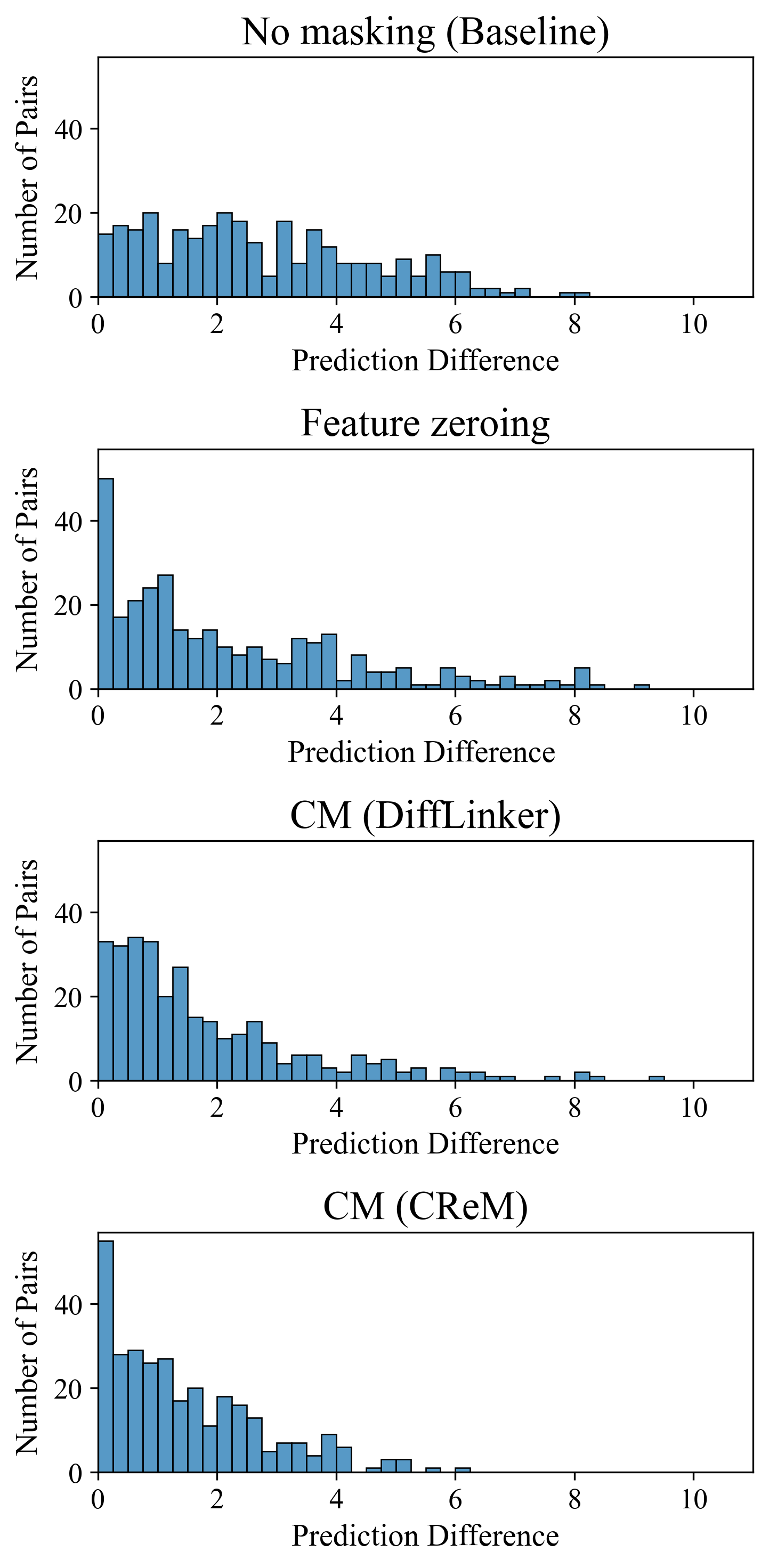}
  \caption{Single Anchor}
  \label{fig:sub1}
\end{subfigure}%
\begin{subfigure}{.5\textwidth}
  \centering
  \includegraphics[width=.9\linewidth]{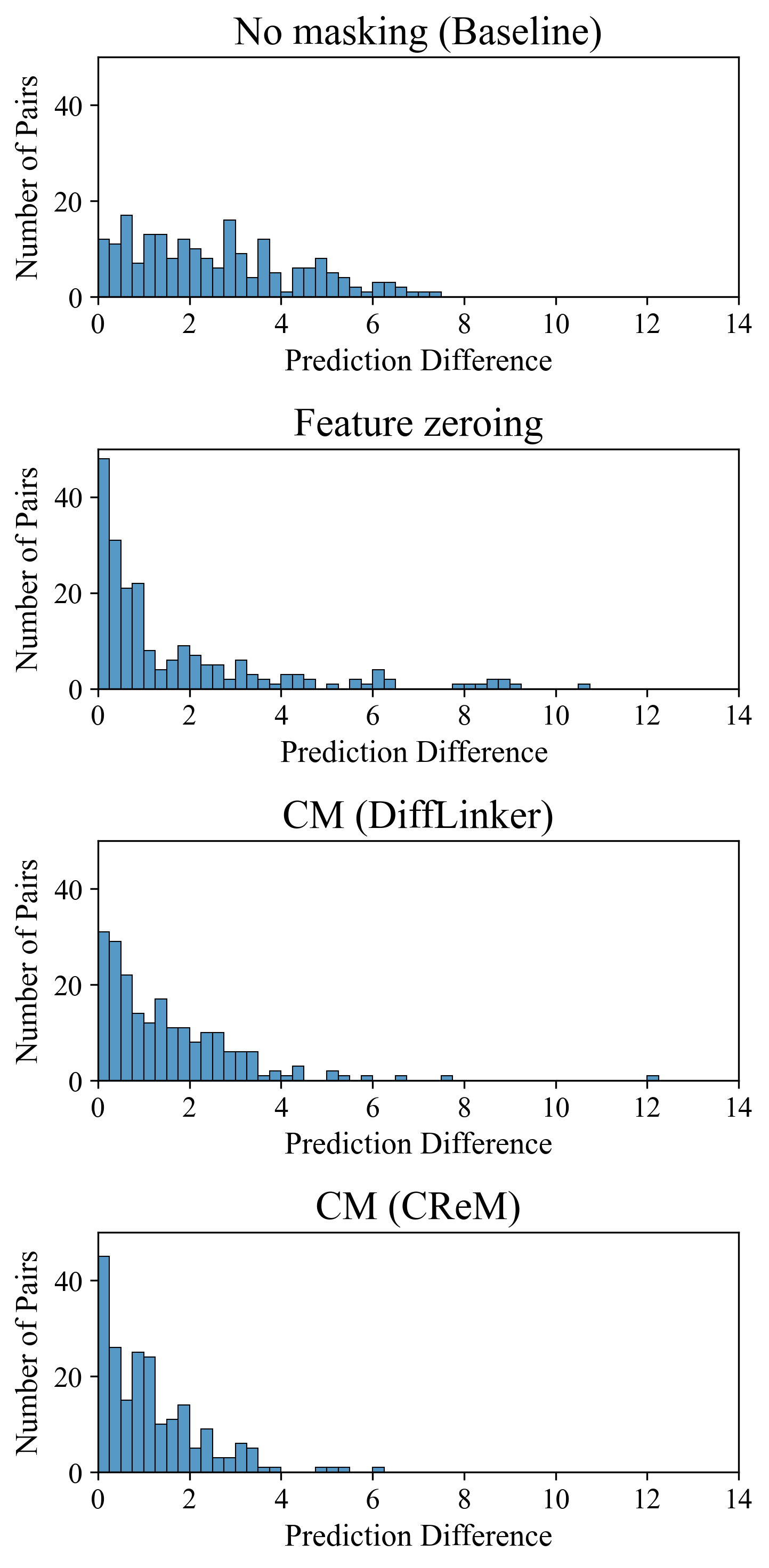}
  \caption{Multiple Anchors}
  \label{fig:sub2}
\end{subfigure}
\caption{\textbf{Histograms of prediction differences for masked molecular pairs with (a) single anchor and (b) multiple anchors connecting the uncommon and common parts.} Ideally, masking the uncommon part of a molecule should result in a prediction difference of zero. Feature zeroing significantly increases the number of pairs with prediction differences close to zero compared to the baseline. However, the distribution also shows a long right tail, indicating that in some cases it even increases the prediction difference. CM (DiffLinker) results in smaller and more stable prediction differences than Feature zeroing, with most samples clustered near zero. CM (CReM) yields the highest number of perfectly masked samples (i.e., prediction difference close to zero) and does not exhibit the same long-tailed distribution as Feature zeroing.}
\label{fig:histograms}
\end{figure*}

Figure~\ref{fig:histograms} presents histograms of prediction differences for masked molecular pairs with (a) a single anchor and (b) multiple anchors connecting the uncommon parts to the common substructure.

\section{C. Comparison of counterfactual examples} 
\begin{figure*}
  \centering
  \includegraphics[width=0.725\linewidth]{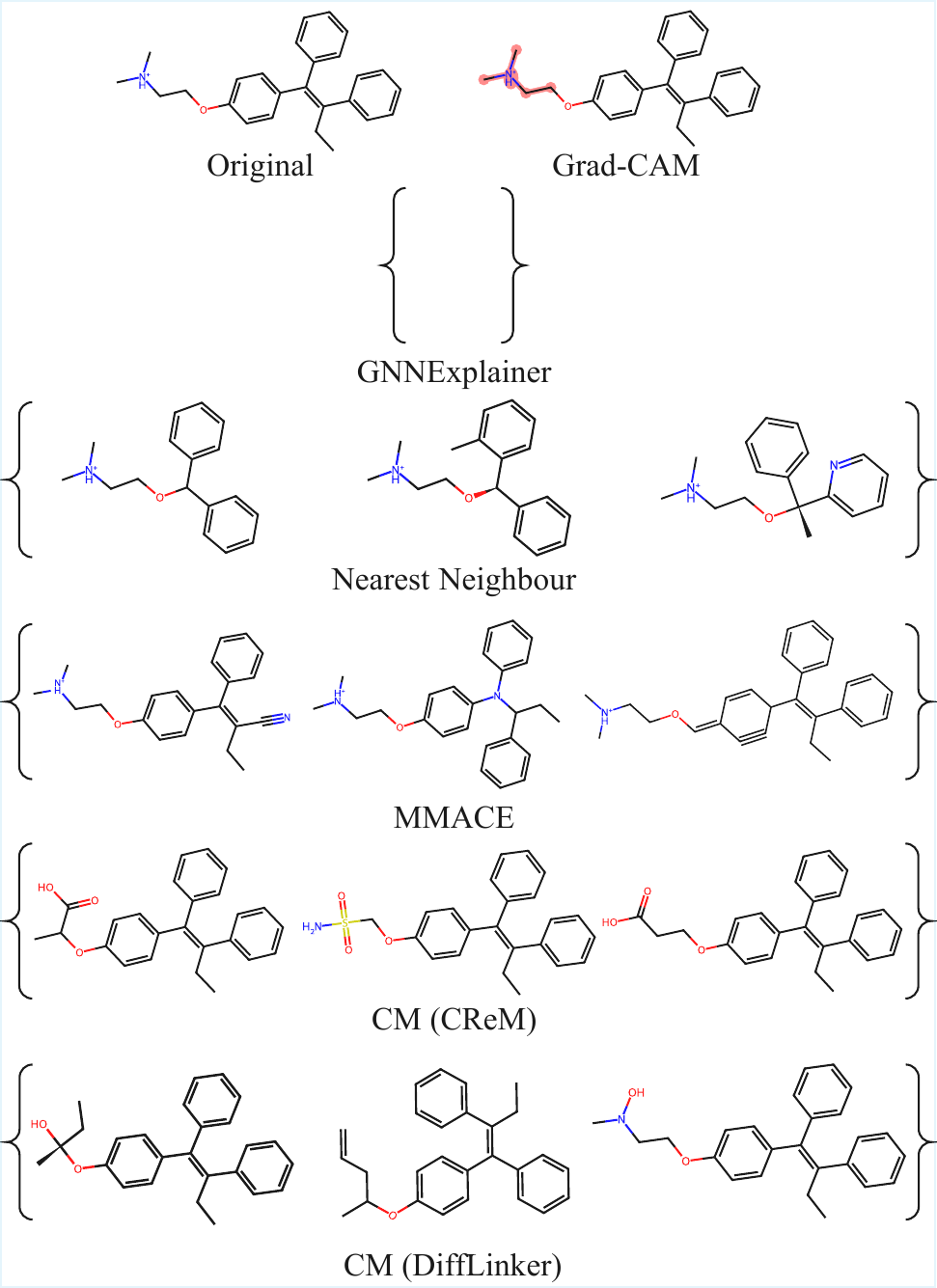}
  \caption{\textbf{Comparison of counterfactual examples generated by tested counterfactuals generation methods.} The initial model prediction classified the original molecule as an hERG blocker. No counterfactual molecule was generated using GNNExplainer, because the atoms it identified for removal caused the molecule to fragment into disconnected parts. Such molecules are considered invalid and discarded. The Nearest Neighbors are molecules from the hERG training dataset that the model predicted as non-blockers, the opposite class of the original. MMACE often produces structurally complex, hard-to-interpret counterfactuals, e.g., containing a triple bond in a ring. Since the MMACE modifications can be applied anywhere across the molecule, drawing clear conclusions from these counterfactuals is often challenging. In contrast, CM (CReM) and CM (DiffLinker) generate chemically plausible and localized modifications by altering only the fragment highlighted by the factual explanation method (here, Grad-CAM). Because these changes are local, the counterfactuals generated by CM methods are generally more interpretable. In five out of six examples, they replace a positively charged tertiary amine with neutral groups, potentially reducing the risk of hERG inhibition by limiting interactions with aromatic amino acids. In the last example, DiffLinker replaces the charged moiety with a non-charged one, possibly achieving a similar effect.}
\label{fig:examples_counterfactuals_herg}
\end{figure*}

\begin{figure*}
  \centering
  \includegraphics[width=0.725\linewidth]{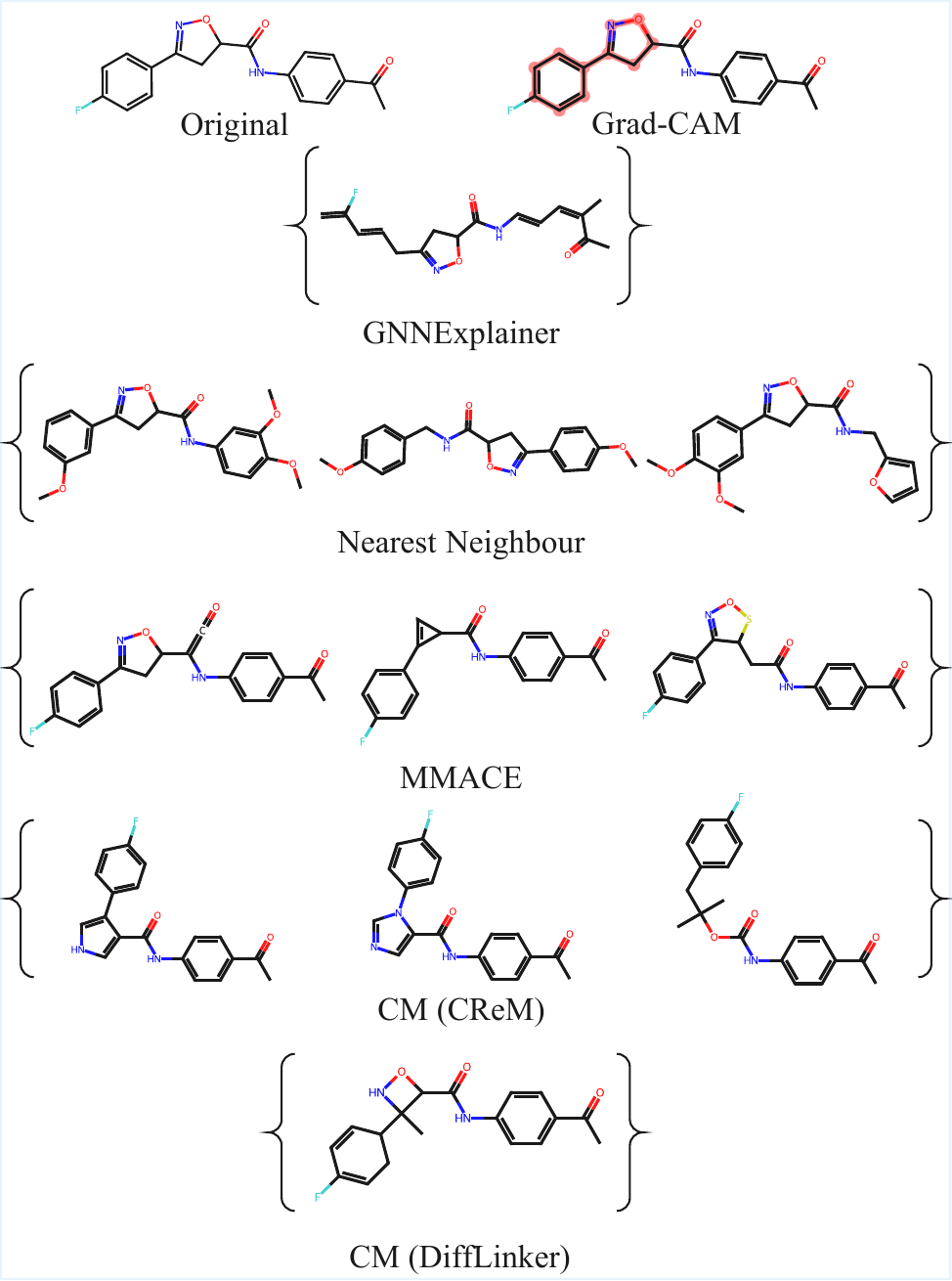}
  \caption{\textbf{Comparison of counterfactual examples generated by tested counterfactual generation methods.} The initial model prediction indicated that the original molecule is \emph{not} an inhibitor of CYP3A4. }  
\label{fig:examples_counterfactuals_cyp3a4}
\end{figure*}

Figure~\ref{fig:examples_counterfactuals_herg} presents generated counterfactuals for a molecule that is an hERG blocker, and Figure~\ref{fig:examples_counterfactuals_cyp3a4} shows generated counterfactuals for a molecule that is not an inhibitor of CYP3A4. 

\section{D. Detailed benchmark results}

The consistency of masking methods and explanation methods, split by model architecture, is shown in Table~\ref{tab:expl-cf-benchmark}.

\begin{table*}
\caption{\textbf{Consistency of masking methods and explanation methods -- split by model architecture}. Model architecture color keys:
    \textcolor{green!18!gray}{GAT},
    \textcolor{red!35!gray}{ResidualGIN},
    \textcolor{cyan!45!gray}{EdgeGIN},
    \textcolor{gray!79}{LargeGIN},
    \textcolor{gray!61}{MediumGIN},
    \textcolor{gray!43}{SmallGIN}.
Corresponds to Table 3 in the main text.
}
\label{tab:expl-cf-benchmark}
\centering
\resizebox{1.0\textwidth}{!}{%
  \input{figures/appendix/table-appendix-d.tex}
}
\end{table*}

\section{E. Qualitative examples}

\begin{figure*}
\includegraphics[width=1.0\textwidth]{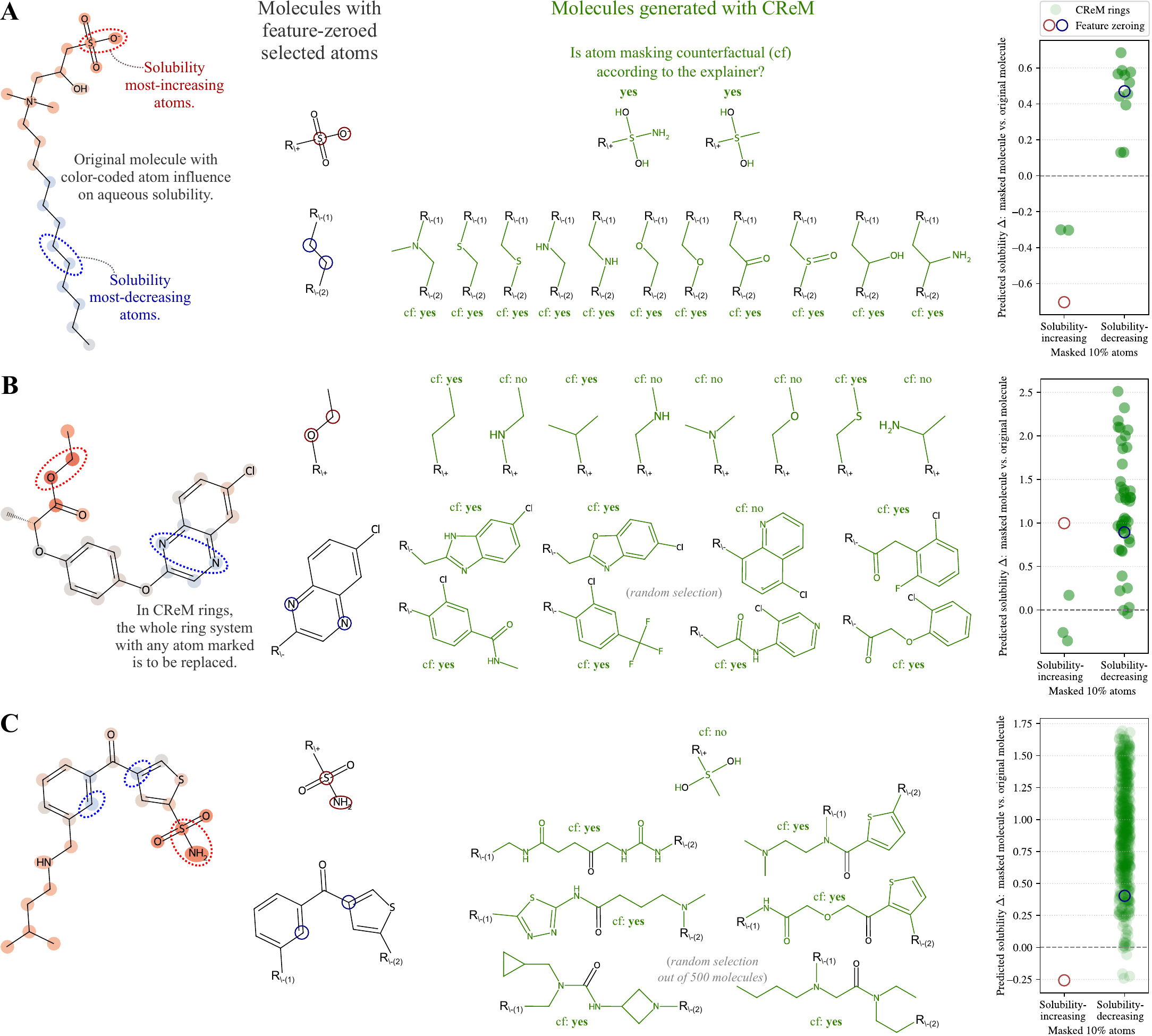}
\caption{
\textbf{Examples illustrating explainer-derived atom contributions to aqueous solubility, alongside corresponding generatively masked molecular variants}.
In each panel, the left-hand structure is a test-set molecule, with color-coded atom-wise influence on solubility: red hues indicate atoms contributing positively (increasing solubility), whereas blue hues indicate negative contributions (reducing solubility).
Right-hand side plots show the impact of feature zeroing (rings) and counterfactual masking with CReM (green disks) on predicted molecule solubility.
(\textbf{A}) Both feature zeroing and counterfactual masking consistently induce changes in predicted solubility aligned with the expected direction of influence.
All CReM-generated molecular variants were identified by the explainer as counterfactuals.
The limited selection of the atoms follows the imposed rule of selecting 10\% of atoms among those most increasing/decreasing the property.
(\textbf{B}) In this example, some CReM-generated molecules were excluded from the analysis due to their classification as non-counterfactuals. Within the CReM rings method, entire ring systems with any atoms deemed highly influential are substituted.
A small subset of generated counterfactuals resulted in changes to the predicted molecular property that did not align with the expected direction of influence.
Additionally, feature zeroing of atoms identified as strongly decreasing solubility failed to produce the anticipated increase in solubility.
This may, in part, be attributed to the masking of (slightly electronegative) nitrogen atoms within the aromatic ring system, which may not serve as effective representatives of the aromatic ring context.
(\textbf{C}) This example showcases the capability of the CReM rings method to replace multiple ring systems simultaneously, the occurrence of extreme imbalance between two pools of generated counterfactual candidates, and the structural diversity of generated molecular variants.
}
\label{fig:explanation-examples}
\end{figure*}

Examples illustrating explainer-derived atom contributions to aqueous solubility, alongside corresponding
generatively masked molecular variants, are shown in Figure~\ref{fig:explanation-examples}.

